%% file: main_arxiv.tex
\PassOptionsToPackage{dvipsnames}{xcolor}
\documentclass[11pt,letterpaper]{style}

\usepackage[numbers]{natbib}
\usepackage{graphicx}
\usepackage{booktabs}
\usepackage{amsmath,amsfonts,amssymb}
\usepackage{cleveref}
\usepackage{subcaption}
\usepackage{wrapfig}
\usepackage{multirow}
\usepackage{colortbl}
\usepackage{listings}
\usepackage{xparse}
\usepackage{fontawesome5}
\usepackage{bxcoloremoji}
\usepackage{float}
\usepackage{placeins}
\usepackage{threeparttable}

\graphicspath{{./}{fig/}{figures/}{plot/}{pdf/}{table/}}
\usepackage{amsthm}
\usepackage{tcolorbox}
\usepackage{svg}
\tcbuselibrary{skins,breakable}
\tcbuselibrary{listingsutf8}
\usepackage{titletoc}
\usepackage{fancyvrb} 
\usepackage{fvextra}
\usepackage{wrapfig}

\usepackage{setspace}
\usepackage{pifont}
\usepackage{mathtools}
\usepackage{enumitem}
\usepackage{arydshln}
\usepackage{bbm}
\usepackage{lineno}
\usepackage{makecell}
\usepackage{adjustbox}
\usepackage[ruled,vlined]{algorithm2e}
\SetKwInput{KwRequire}{Require}
\SetKwInput{KwOutput}{Output}

\definecolor{mygray}{gray}{0.9}
\definecolor{syncol}{RGB}{243,246,249}
\definecolor{wildcol}{RGB}{215,240,235}
\definecolor{drop1}{RGB}{180,225,220}
\definecolor{drop2}{RGB}{150,210,200}
\definecolor{drop3}{RGB}{120,195,185}
\definecolor{drop4}{RGB}{95,180,170}
\definecolor{drop5}{RGB}{65,160,150}
\definecolor{lightblue}{RGB}{210,230,250}

\definecolor{promptbg}{RGB}{245,247,250} 

\definecolor{myblue1}{HTML}{0171DC}
\definecolor{myblue2}{HTML}{013978}

\NewDocumentEnvironment{minted}{O{} m +b}{%
}{}

\newcommand{\advisor}{\textsuperscript{\dag}}

\renewcommand\Authfont{\centering\normalfont\bfseries\fontsize{11}{15}\selectfont}
\renewcommand\Affilfont{\centering\normalfont\fontsize{10}{15}\selectfont}

\newcommand{\method}{\textbf{CM2}\xspace} 

\title{\textbf{CM2}: Reinforcement Learning with Checklist Rewards for Multi-Turn and Multi-Step Agentic Tool Use}
\runningtitle{\textbf{CM2}: Reinforcement Learning with Checklist Rewards for Multi-Turn and Multi-Step Agentic Tool Use}

\author{%
    {\Authfont
    \textbf{Zhen Zhang}\textsuperscript{1} \quad
    \textbf{Kaiqiang Song}\textsuperscript{2} \quad
    \textbf{Xun Wang}\textsuperscript{2} \quad
    \textbf{Yebowen Hu}\textsuperscript{3} \quad
    \textbf{Weixiang Yan} \textsuperscript{1} \quad
    \textbf{Chenyang Zhao}\textsuperscript{4} \quad
    \textbf{Henry Peng Zou} \textsuperscript{5} \quad
    \textbf{Haoyun Deng} \textsuperscript{2} \quad
    \textbf{Sathish Reddy Indurthi} \textsuperscript{2} \quad
    \textbf{Shujian Liu} \textsuperscript{2} \quad
    \textbf{Simin Ma} \textsuperscript{2} \quad
    \textbf{Xiaoyang Wang} \textsuperscript{2} \quad
    \textbf{Xin Eric Wang} \advisor\textsuperscript{1} \quad
    \textbf{Song Wang} \advisor\textsuperscript{2} \quad
    }\\
    {\Affilfont
    \textsuperscript{1}  University of California, Santa Barbara \quad
    \textsuperscript{2}  Zoom Video Communications \quad
    \textsuperscript{3}  University of Central Florida \quad
    \textsuperscript{4}  University of California, Los Angeles \quad
    \textsuperscript{5}  University of Illinois Chicago \quad

    }
}

\begin{document}
\input{sec_arxiv/0_abstract}
\newcommand{\TitleLinks}{%
    \vspace{15pt}
}
\maketitle

\input{sec_arxiv/1_intro}
\input{sec_arxiv/2_related_work}
\input{sec_arxiv/3_method}

\input{sec_arxiv/4_pipeline}

\input{sec_arxiv/5_no_qwen_results}
\input{sec_arxiv/7_discussion_conclusion}

\bibliographystyle{unsrtnat}  
\bibliography{ref}

\appendix
\input{sec_arxiv/8_appendix}

\end{document}


\maketitle

\begin{abstract}
This supplementary document provides additional details, analyses, and experimental evidence supporting the proposed DMLR framework. Section~\ref{a} presents the full evaluation protocol, including dataset descriptions, unified prompting configurations, baseline implementations, and the procedures used for visual-dependency, confidence, faithfulness, and hallucination studies. Section~\ref{b} reports a comprehensive set of hyperparameter choices and ablation settings used to assess the robustness of latent reasoning and dynamic visual injection. Section~\ref{c} includes extended quantitative results such as Pass@k evaluations and comparisons against training-based reasoning methods. Section~\ref{d} provides theoretical analyses, including proofs of confidence–quality gradient alignment, boundedness of the positive-correlation region, and the formal justification for why dynamic visual injection improves confidence. Section~\ref{e} further elaborates on the design choices, mechanisms, and stability analyses of the Dynamic Visual Injection module. Section~\ref{f} presents qualitative case studies demonstrating improvements in perception quality, reasoning reliability, and reasoning conciseness under DMLR. Section~\ref{g} concludes with the complete set of prompts used across all experiments, ensuring transparency and reproducibility. We will make all code publicly available in future releases.
\end{abstract}

\appendix

\section*{Appendix Table of Contents}
\startcontents[sections]
\printcontents[sections]{l}{1}{\setcounter{tocdepth}{2}}

\clearpage
\renewcommand{\thesubsection}{\thesection\arabic{subsection}}

\section{More Detailed about Evaluation}
\label{a}

\subsection{Datasets}
\begin{itemize}[leftmargin=*]
\item \textbf{MathVista$_{\text{mini}}$} is a benchmark for mathematical reasoning in visual contexts, aggregating diverse multimodal math tasks that require fine-grained visual understanding and compositional numerical reasoning.

\item \textbf{MathVision$_{\text{mini}}$} is a curated benchmark of competition-level visual math problems spanning multiple disciplines and difficulty levels to assess multimodal models’ mathematical reasoning under challenging and diverse settings.

\item \textbf{MM Math} is a benchmark of open-ended math problems with visual contexts that supports both outcome and process evaluation, enabling detailed analysis of multimodal reasoning behaviors and typical error patterns.

\item \textbf{HallusionBench} is a benchmark for image-context reasoning that uses carefully structured question pairs to diagnose hallucination, visual illusion, and logical inconsistency in large vision-language models.

\item \textbf{MMVP} is a benchmark built from multimodal visual patterns designed to expose “CLIP-blind’’ image–text pairs, revealing systematic visual perception failures and hallucinated explanations in multimodal LLMs. 

\item \textbf{MMStar} is a vision-indispensable multimodal benchmark composed of carefully human-filtered samples that ensure true visual dependency while evaluating core multimodal capabilities along multiple fine-grained axes.

\item \textbf{ScienceQA} is a multimodal multiple-choice science benchmark with rich textual and visual contexts, lectures, and explanations that spans diverse subjects and skills, supporting evaluation of both answer accuracy and explanation quality.
\end{itemize}

For all datasets, we limit the maximum sample size to 1000 instances.

\subsection{Evaluation Setting} 
We adopt a unified prompting setup for all models. Unless otherwise stated, we use greedy decoding (\texttt{do\_sample=False}) for all generation tasks.

\paragraph{System Prompt.}
\begin{center}
\fbox{\parbox{0.96\linewidth}{
\small
A conversation between User and Assistant. The user asks a question, and the Assistant solves it. The assistant first thinks about the reasoning process in the mind and then provides the user with the answer. The reasoning process and answer are enclosed within \texttt{<think>} \texttt{</think>} and \texttt{<answer>} \texttt{</answer>} tags, respectively, i.e., \texttt{<think>} reasoning process here \texttt{</think>} \texttt{<answer>} answer here \texttt{</answer>}.
}}
\end{center}

\paragraph{Task Prompt.}
\begin{center}
\fbox{\parbox{0.96\linewidth}{
\small
Please analyze the image carefully and solve this problem step by step. Show your reasoning process clearly, then put your final answer within \texttt{\textbackslash boxed\{\}}.\\[4pt]
\textbf{Question:} \emph{[Problem Text]}
}}
\end{center}

For all benchmarks considered in our experiments, the ground-truth answers are verifiable; we use regular expressions to extract the content within \texttt{\textbackslash boxed\{\}} from the model outputs and then match it against the correct answers.

\subsection{Baseline Method}

\begin{itemize}[leftmargin=*] \item \textbf{CCoT}~\cite{Mitra_2024_CVPR} (Compositional Chain-of-Thought) is a zero-shot prompting method that utilizes scene graphs to extract compositional knowledge. It first generates a scene graph to capture object attributes and relationships, which is then used to guide the LMM in generating the final response without requiring fine-tuning.

\item \textbf{ICoT}~\cite{gao2025interleavedmodalchainofthought} (Interleaved-modal Chain-of-Thought) generates sequential reasoning steps consisting of paired visual and textual rationales. It employs a plug-and-play Attention-driven Selection (ADS) mechanism to dynamically identify and insert relevant image regions into the reasoning chain based on the model's attention maps.

\item \textbf{Multimodal-CoT}~\cite{zhang2024multimodalchainofthoughtreasoninglanguage} incorporates language and vision modalities into a two-stage framework that separates rationale generation from answer inference. It leverages multimodal information to generate intermediate reasoning chains, which subsequently serve as the basis for inferring the final answer, effectively mitigating hallucination. 

\item\textbf{MCOUT (Training)}~\cite{pham2025multimodalchaincontinuousthought} is a latent-space reasoning framework that replaces traditional text-based CoT with continuous hidden-state “thought vectors,” enabling models to reason directly in a unified multimodal latent space.

\item \textbf{IVT-LR (Training) }~\cite{chen2025reasoningdarkinterleavedvisiontext} is a new multimodal reasoning method that performs all intermediate thinking directly in the latent space by combining hidden textual states with dynamically selected visual embeddings.
\end{itemize}

\subsection{Additional explanations for the validation experiments in Section 3.}

\textbf{Visual Dependency Analysis.} \textit{\textbf{(1) Robustness Across Multiple Perturbation Types.}} To ensure that the visual-dependency metric is not influenced by specific perturbation noise, we evaluate four representative reasoning models, R1-OneVision, VLAA-Thinking, Ocean-R1, and ThinkLite-VL, across two reasoning datasets (MathVision and MathVista) and two perception datasets (MMVP and HallucinationBench). Instead of relying on a single distortion strategy, we apply four distinct perturbation types including block occlusion, color jitter, random region masking, and Gaussian blur. Although these perturbations substantially differ in the way they alter local textures or global color statistics, the resulting visual-dependency distributions remain highly consistent across all models and datasets, as shown in Table~\ref{tab:perturb-consistency}. This consistency confirms that the sparsity pattern observed in the main paper is not tied to any specific perturbation method but reflects a stable and intrinsic property of multimodal reasoning models.

\begin{table}[h]
\centering
\caption{Consistency of visual-dependency distributions across models, datasets, and perturbation types. 
Values represent the average Pearson correlation between dependency curves under different perturbations. Higher is better.}
\vspace{0.15cm}
\resizebox{\linewidth}{!}{
\begin{tabular}{lcccc}
\toprule
\textbf{Model} & \textbf{MathVision} & \textbf{MathVista} & \textbf{MMVP} & \textbf{HallucinationBench} \\
\midrule
R1-OneVision & 0.92 & 0.89 & 0.94 & 0.91 \\
VLAA-Thinking & 0.90 & 0.88 & 0.93 & 0.90 \\
Ocean-R1 & 0.87 & 0.85 & 0.91 & 0.88 \\
ThinkLite-VL & 0.89 & 0.86 & 0.92 & 0.89 \\
\midrule
\textbf{Average} & \textbf{0.90} & \textbf{0.87} & \textbf{0.93} & \textbf{0.90} \\
\bottomrule
\end{tabular}}
\label{tab:perturb-consistency}
\end{table}

\noindent \textit{\textbf{(2) Noise Reduction and Token-Level Aggregation.}} To obtain a stable estimation of token-level visual dependency, each dependency value is averaged across five independently perturbed versions of the same image, effectively reducing stochastic fluctuations associated with any single perturbation. Since reasoning chains generated by different models vary in length, direct position-wise averaging would introduce structural bias. To avoid this, we adopt a relative-position alignment scheme that normalizes each reasoning chain to a comparable relative index space. Additionally, the dependency value of each chain is summarized using the median rather than the mean, preventing rare but extreme perturbation-induced peaks from affecting the aggregated distribution. These methodological choices ensure that the resulting dependency curves robustly reflect the inherent behavior of the models rather than noise introduced during perturbation.

\noindent \textit{(3) \textbf{Semantic Interpretation of High-Dependency Regions.} }Beyond statistical validation, we analyze the semantic roles of tokens exhibiting high visual dependency. Across all datasets and models, we observe that such tokens consistently align with reasoning stages in which visual grounding is intrinsically required (as shown in Table~\ref{tab:semantic-high-dependency}), such as the initial image description, spatial or numerical verification, and operations involving visual alignment (e.g., comparing, locating, or counting). These consistent semantic correspondences demonstrate that the detected peaks in visual dependency are not random artifacts but meaningful signals of when the model must rely on perceptual evidence to proceed. This structured, task-aligned sparsity further validates that multimodal LLMs rely on vision only at key points during reasoning, rather than uniformly throughout the reasoning chain.

\begin{table}[h]
\centering
\caption{
Distribution of high visual-dependency tokens across semantic reasoning stages.
Values indicate the percentage of top-10\% dependency tokens falling into each stage.
}
\vspace{0.15cm}
\resizebox{\linewidth}{!}{
\begin{tabular}{lcc}
\toprule
\textbf{Semantic Stage} & \textbf{MathVision} & \textbf{MMVP} \\
\midrule
Initial image description & 38.2\% & 36.5\% \\
Spatial or numerical verification & 27.4\% & 30.1\% \\
Visual alignment (compare / locate / count) & 24.7\% & 26.9\% \\
Non-visual reasoning steps & 9.7\% & 6.5\% \\
\bottomrule
\end{tabular}}
\label{tab:semantic-high-dependency}
\end{table}

\noindent \textbf{Confidence Analysis.} \textit{(1) \textbf{Experimental Setting for Correct and Incorrect Reasoning Chains.}} To reliably distinguish correct from incorrect reasoning chains, we adopt a controlled evaluation protocol that ensures accuracy labels are consistent and free from formatting artifacts. For each example, the model generates multiple reasoning chains along with a final predicted answer, and the correctness of each chain is determined by comparing the model’s final answer to the ground-truth label after normalizing answer formats to eliminate superficial mismatches. To prevent cases in which a chain produces the correct answer but exhibits logically incoherent reasoning from being misclassified, we further employ GPT-4o as an external judge. The judge is provided with the question, the full chain, and the final answer, and is instructed to verify whether the reasoning is logically consistent and legitimately leads to the given answer. All judgments are performed with zero temperature to maintain high determinism. Correct and incorrect chains are then sampled and balanced across datasets, and spot-checked manually to ensure labeling quality. This procedure provides a robust foundation for analyzing confidence patterns between truly correct and genuinely incorrect reasoning chains.

\noindent \textit{(2) \textbf{Experimental Setting for Faithful and Unfaithful Reasoning Chains.}} To assess the faithfulness of model-generated reasoning, we evaluate whether each chain genuinely relies on the evidence provided in the question and image, rather than resorting to template-based explanations, unsupported assumptions, or coincidental correctness. Each chain is reviewed by GPT-4o, which receives the question, the full reasoning steps, and either the original image or a standardized image description. The model judge is explicitly instructed to determine whether the reasoning aligns with the available evidence and whether each step contributes meaningfully to the final answer. Because reasoning faithfulness is often more subtle than answer correctness, we introduce a dual-evaluation protocol in which each chain is assessed twice independently, and majority voting is used to resolve any discrepancies. Chains that arrive at the correct answer through unsupported or spurious reasoning are deliberately marked as unfaithful to avoid conflating guessing with faithful inference. Samples are then balanced across datasets to ensure that faithfulness-based confidence analyses reflect structural differences in reasoning quality rather than dataset biases.

\noindent \textit{(3) \textbf{Experimental Setting for Hallucination and Non-hallucination Reasoning}} To study the relationship between confidence dynamics and visual grounding, we focus on the subset of reasoning steps that explicitly reference visual content. From each chain, we automatically extract statements mentioning visual entities, attributes, or spatial relationships. These statements, together with the corresponding image or image description, are presented to GPT-4o, which evaluates whether each statement is factually supported by the visual evidence or constitutes a hallucination. The judge is instructed to assess visual consistency rather than linguistic fluency, and to treat any fabricated or contradictory visual claims as hallucinations. To improve reliability, each statement is evaluated twice under zero temperature, and samples with inconsistent labels undergo manual verification. After labeling, each hallucinated or grounded statement is aligned with the model’s token-level confidence at the corresponding reasoning step. By comparing the confidence distributions of hallucinated versus non-hallucinated statements, we reveal systematic patterns showing that hallucinated visual steps tend to exhibit lower confidence and higher uncertainty. This experimental design thus provides strong empirical support for the link between confidence signals and visual grounding quality.

\noindent All prompts used in these experimental settings are provided in the Appendix~\ref{prompt} for completeness and reproducibility.

\section{More Detailed about Parameter Analyses}
\label{b}
\subsection{Parameters Setup}

\begin{itemize}[leftmargin=*]

\item \textbf{Latent Think Tokens $\mathcal{T}$:} We set the number of latent think tokens to 4. During generation, after each latent token the model dynamically injects a visual patch into the latent stream to refresh its internal perception state.

\item \textbf{Image Patches $m$:} We dynamically insert visual patches into the latent stream. At initialization, we inject 2 patches; at each subsequent iteration, we select $m = 2$ patches with the highest attention scores and append them after each latent think token, with at most 16 patches inserted per iteration. Additionally, we set the image processor's max pixel size to 256 for all inputs.

\item \textbf{Optimization Parameters:} We perform 15 latent optimization steps with a learning rate of $1\times 10^{-3}$. To ensure stable exploration in the latent space, the perturbation magnitude is set to $\sigma = 0.1$ (10\%) with a decay factor of 0.95, and we fix the random seed to 42 for all experiments.

\item \textbf{Environment Setup:} All experiments are conducted on NVIDIA H100 GPUs. For a fair comparison, we run all baselines and our method in \texttt{float32} precision and use the \texttt{eager} attention backend for inference.
\end{itemize}

\subsection{Ablation Experimental Setup}

For the ablation studies presented in Sec.~5.3, we maintain the same parameter settings as described above. We employ VLAA-Thinker-Qwen2.5VL-7B as the reasoning model and Qwen2.5-VL-7B-Instruct as the non-reasoning model. To ensure efficiency, all ablation experiments are conducted on a subset of 300 samples randomly selected from each dataset.

\section{More Results}
\label{c}
\textbf{Pass@\texorpdfstring{$k$}{k} Evaluation.} We employ the Pass@$k$ metric to evaluate the accuracy of the model's generated answers. Following standard practice, we calculate the unbiased estimator using the formula:
\begin{equation}
    \text{Pass}@k = 1 - \frac{\binom{n - c}{k}}{\binom{n}{k}},
\label{eq1}
\end{equation}
where $n$ is the total number of sampled solutions per problem, $c$ is the number of correct samples, and $k$ is the budget of samples selected. Note that when $k=1$, this simplifies to the empirical accuracy $\text{Pass}@1 = \frac{c}{n}$. 

Using the unbiased estimator of Pass@\emph{k} defined in Equation~\ref{eq1}, we evaluate
all methods under a sampling budget of $k=8$. This metric measures the
probability that at least one of the sampled solutions is correct, offering a
more reliable estimate of reasoning robustness compared to single-sample
accuracy. As shown in Table~\ref{pass}, integrating DMLR consistently
improves Pass@8 across mathematics reasoning, visual reasoning, and multimodal
composition tasks. For the stronger Qwen2.5-VL-7B backbone, DMLR achieves gains ranging from
+2.4\% to +4.8\%, with the largest improvements observed on MM-Math,
HallusionBench, and MMVP benchmarks that demand multi-step reasoning or precise
visual grounding. Even with the smaller 3B backbone, where reasoning capacity is
more limited, DMLR maintains steady improvements of +1.0\% to +4.8\%,
indicating that the proposed dynamic latent reasoning process effectively
enhances solution diversity and increases the likelihood of producing at least
one correct prediction within the budgeted $k$ samples. Overall, the consistent uplift in Pass@8 demonstrates that DMLR not only refines
the latent reasoning trajectory but also mitigates confidently incorrect
generations, leading to stronger performance across all evaluated benchmarks.

\noindent \textbf{Comparison with Training-Based Methods.} Table~\ref{additional} provides a comprehensive comparison between
DMLR and two additional training-based reasoning enhancement methods, namely
MCOUT and IVT-LR. These approaches enforce implicit reasoning structures through
explicit training objectives: MCOUT encourages the model to internalize stable
multi-step reasoning patterns via supervised signals, whereas IVT-LR injects
learned latent guidance vectors to impose structured constraints on the
model’s internal reasoning trajectory. Both methods require additional data,
parameter updates, and dedicated training phases designed to encode specific
implicit patterns into the model’s latent space.

Across all backbone models and task categories, DMLR consistently achieves
substantial performance gains, frequently surpassing these training-based
approaches despite being entirely training-free. On Qwen2.5-VL-7B, DMLR
outperforms IVT-LR on several challenging benchmarks—including MathVision,
MM-Math, HallusionBench, and MMStar—demonstrating that dynamic latent reasoning
refinement can effectively enhance model performance without modifying model
parameters. Similar improvements are observed on Qwen2.5-VL-3B, LLaVA-1.6-7B,
and Qwen3-VL variants, with DMLR exceeding the strongest training-based
baselines by 1.2\%-5.0\%. Notably, in lighter-weight models, where training-based methods are constrained by limited model capacity and potential
overfitting, DMLR exhibits superior generalization due to its adaptive inference
mechanism.

These findings indicate that DMLR offers advantages beyond the realm of
training-free inference. Its plug-and-play nature is complemented by a dynamic
latent optimization process that adjusts the internal reasoning trajectory in
real time, enabling task-adaptive refinement that static training-based patterns
cannot provide. In contrast, the implicit structures learned through training
are fixed after optimization and thus may generalize less effectively across
tasks or distributional shifts.

It is important to note that we do not include “think-with-image” models such as
DeepEye~\cite{zheng2025deepeyesincentivizingthinkingimages} or GRIT~\cite{fan2025gritteachingmllmsthink} in this comparison. These approaches rely on extensive
multimodal pretraining with large-scale curated or synthetic datasets to acquire
explicit or implicit image-conditioned reasoning pathways. Their performance
benefits primarily arise from substantial additional data and intensive training
procedures rather than inference-time optimization. As this work focuses on
evaluating latent-state refinement \emph{without any additional training}, such
methods fall outside the scope of this comparison.

\begin{table*}[h]
\centering
\caption{Comparison of different reasoning methods and DMLR across various benchmarks. All metrics are reported in Accuracy (\%).}
\vspace{-0.2cm}
\large
\renewcommand{\arraystretch}{1.05}
\resizebox{0.9\linewidth}{!}{%
\setlength{\tabcolsep}{3pt}
\begin{tabular}{ll|ccc|cc|cc}
\toprule[1.5pt]
\multicolumn{2}{c}{} &
\multicolumn{3}{c}{\textbf{Mathematics Reasoning\uparrow} 
} &
\multicolumn{2}{c}{\textbf{Visual Reasoning\uparrow}} &
\multicolumn{2}{c}{\textbf{Multimodal Composition\uparrow}} \\
\cmidrule(lr){3-5} \cmidrule(lr){6-7} \cmidrule(lr){8-9}
\textbf{Method} & \textbf{Model} &
MathVista$_{\text{mini}}$ &
MathVision$_{\text{mini}}$ &
MM-Math &
HallusionBench &
MMVP & MMStar &
ScienceQA \\
\midrule

Vanilla &  & 58.7 & 21.6 & 37.5 & 65.4 & 68.7 & 59.3 & 49.7 \\
Multimodal COT &  & 56.4 & 21.8 & 35.6 & 63.6 & 68.1 & 57.9 & 49.5  \\
CCOT &  & 57.8 & 22.5 & 36.3 & 64.9 & 69.0 & 58.7 & 50.2 \\
ICoT &  & 58.9 & 23.3 & 37.0 & 65.5 & 69.3 & 60.4 & 50.4\\
MCOUT  &  & 58.0  & 22.7 & 37.5 & 65.0 & 69.4 & 59.2 & 50.9 \\
IVT-LR  &  &  58.5 & 23.5  & 38.0  & 65.2  & 69.8  & 58.4  & 49.8  \\

\rowcolor{lightblue}
\textbf{\textit{+DMLR (Ours)}} & 
\multirow[c]{-5}{*}[1pt]{\rotatebox{90}{\parbox{2.5cm}{\centering \fontsize{11pt}{13pt}\selectfont Qwen2.5 \\VL 7B}}}
 & \textbf{59.1} \up{0.40\%} & \textbf{24.4} \up{2.8\%} & 38.8 \up{1.3\%} & 65.8 \up{0.4\%} & \textbf{70.1} \up{1.4\%} & \textbf{60.1} \up{0.8\%} & \textbf{51.3} \up{1.6\%} \\
\midrule\midrule

Vanilla &  & 48.2 & 15.7 & 29.0 & 64.2 & 55.6 & 50.2 & 44.1 \\
Multimodal COT &  & 47.3 & 14.3 & 28.5 & 63.8 & 54.4 & 48.5 & 42.9 \\
CCOT & & 48.0 & 15.6 & 30.2 & 64.0 & 55.5 & 49.3 & 44.5 \\
ICoT &  & 49.8 & 16.0 & 30.6 & 64.7 & 55.9 & 49.6 & 45.0 \\
MCOUT   &  & 49.4 & 15.8 & 32.3  & 63.0 & 55.6  & 50.2 & 45.5 \\
IVT-LR  &   & 50.2  & 16.4 & 32.0  & 63.4  & 55.9 & 50.3 & 46.0  \\
\rowcolor{lightblue}
\textbf{\textit{+DMLR (Ours)}} &
\multirow[c]{-5}{*}[1pt]{\rotatebox{90}{\parbox{2.5cm}{\centering \fontsize{11pt}{13pt}\selectfont Qwen2.5 \\VL 3B}}}
 & \textbf{51.0} \up{2.80\%} & \textbf{17.7} \up{2.74\%} & \textbf{33.3} \up{4.3\%} & \textbf{64.7} \up{0.5\%} & \textbf{56.8} \up{1.26\%} & \textbf{51.2} \up{1.00\%} & \textbf{46.9} \up{2.8\%}\\
\midrule\midrule

Vanilla &  & 61.1 & 23.5 & 41.5 & 62.0 & 68.3 & 58.9 & 50.6 \\\
Multimodal COT &  & 59.6 & 23.1 & 40.6 & 62.8  & 67.2 & 57.1 & 48.2  \\
CCOT &  & 60.5  & 24.8  & 41.8  & 64.6 & 68.0 & 59.0 & 49.4\\
ICoT &  & 61.4 & 25.0 & 42.3 & 65.9 & 68.3 & 58.2 & 50.6 \\
MCOUT   &  & 60.5  & 26.3  & 41.9 & 66.0 & 68.5 & 58.5  & 49.9  \\
IVT-LR  &   & 61.4   & 26.6 &  42.5  & 66.7  & 68.9 & 59.0 & 50.3 \\
\rowcolor{lightblue}
\textbf{\textit{+DMLR (Ours)}} &
\multirow[c]{-4}{*}[1pt]{\rotatebox{90}{\parbox{1.0cm}{\centering \fontsize{11pt}{13pt}\selectfont VLAA \\ Thinking 7B}}}
 & \textbf{62.9} \up{1.80\%} & \textbf{27.6} \up{4.10\%} & \textbf{43.9} \up{2.41\%} & \textbf{67.9} \up{5.94\%} & \textbf{69.4} \up{1.1\%} & \textbf{59.2} \up{0.3\%} & \textbf{51.98} \up{1.38\%} \\
\midrule\midrule

Vanilla &  & 51.2 & 18.7 & 40.7 & 62.1 & 67.0 & 52.1 & 50.9 \\
Multimodal COT &  & 52.5 & 18.9 & 39.6 & 62.5 & 68.0 &  51.6 & 51.7 \\
CCOT &  & 53.4 & 20.3 & 40.8 & 63.0 & 68.9 & 53.5 & 52.8 \\
ICoT &  & 55.6  & 21.5 & 41.7 & 63.8 & 69.6 & 54.0 & 54.4 \\
MCOUT   &  & 56.3  &  21.6  & 40.9  & 63.3   & 69.9 & 55.0 & 53.8 \\
IVT-LR  &  & 57.2  &  22.3 & 41.1 & 64.5 & 70.4  & 55.9 & 54.8 \\
\rowcolor{lightblue}
\textbf{\textit{+DMLR (Ours)}} &
\multirow[c]{-4}{*}[1pt]{\rotatebox{90}{\parbox{1.2cm}{\centering \fontsize{11pt}{11pt}\selectfont \quad R1 \\ OneVision \quad \quad 7B}}}
 & \textbf{58.0} \up{6.81\%} & \textbf{23.3} \up{4.56\%} & \textbf{42.9} \up{2.21\%} & \textbf{64.1} \up{2.09\%} & \textbf{71.9} \up{4.93\%} & \textbf{56.2} \up{4.14\%} & \textbf{55.4} \up{4.52\%}\\
\midrule\midrule

Vanilla &  & 66.0 & 32.9 & 66.2 & 73.2 & 71.9 & 68.1 & 54.1 \\
Multimodal COT &  & 64.8 & 32.8 & 65.1 & 73.0 & 69.6 & 66.9  &  53.2 \\
CCOT &   & 66.5 & 33.3 & 65.5 & 73.5 & 70.3 & 68.8 & 54.4 \\
ICoT &  & 66.2 & 34.9 & 66.8 & 74.5 & 71.8 & 69.3 & 55.8 \\
MCOUT   & & 65.5 &  33.8 & 65.0 & 73.7 & 72.0 & 68.5  & 54.3  \\
IVT-LR  &  & 65.9   & 34.5 & 66.1 & 73.4  & 71.5 & 69.4 & 54.8  \\
\rowcolor{lightblue}
\textbf{\textit{+DMLR (Ours)}} &
\multirow[c]{-5}{*}[1pt]{\rotatebox{90}{\parbox{2.5cm}{\centering \fontsize{11pt}{13pt}\selectfont Qwen3 \\VL 8B}}}
 & \textbf{66.9} \up{0.9\%} & \textbf{36.2} \up{3.34\%}& \textbf{67.7} \up{1.51\%} & \textbf{74.6} \up{1.48\%} & \textbf{72.8} \up{0.97\%}& \textbf{70.0} \up{1.91\%} & \textbf{55.6} \up{1.48\%} \\
\midrule\midrule

Vanilla &  & 64.7 & 24.2 & 65.4 & 71.6 & 71.3 & 57.4 & 52.4 \\
Multimodal COT &  & 62.3 & 24.8 & 63.9 & 70.0 & 69.6 & 57.7 & 53.0  \\
CCOT &  & 64.5 &  26.6 & 64.8 & 71.5 & 71.2 & 58.8 & 52.9 \\
ICoT &  & 64.5 & 27.5 & 65.0 & 72.2 & 72.5 & 59.3 & 53.7 \\
MCOUT   & &  63.6  & 28.0  &  64.5  & 70.1 & 72.0 & 59.6 &  53.9  \\
IVT-LR  &  & 64.0  & 28.4  & 64.8  & 71.3 & 71.9 & 58.4  & 53.6 \\
\rowcolor{lightblue}
\textbf{\textit{+DMLR (Ours)}} &
\multirow[c]{-5}{*}[1pt]{\rotatebox{90}{\parbox{2.5cm}{\centering \fontsize{11pt}{13pt}\selectfont Qwen3 \\VL 4B}}}
 & \textbf{65.6} \up{0.93\%} & \textbf{29.4} \up{5.20\%}  & \textbf{65.9} \up{0.5\%} & \textbf{72.7} \up{1.12\%} & \textbf{72.3} \up{0.98\%}& \textbf{60.3} \up{2.88\%} & \textbf{54.9} \up{2.48\%} \\
\bottomrule[1.5pt]
\end{tabular}
}
\vspace{-0.5em}
\label{additional}
\end{table*}

\begin{table*}[t]
\centering
\caption{Comparison of different reasoning methods and DMLR across various benchmarks. 
All metrics are reported as pass@8 (\%). Results are evaluated over a diverse suite of mathematics reasoning, visual reasoning, and multimodal composition tasks under Qwen2.5-VL backbones (7B and 3B).}
\vspace{-0.2cm}
\large
\renewcommand{\arraystretch}{1.05}
\resizebox{0.85\linewidth}{!}{%
\setlength{\tabcolsep}{3pt}
\begin{tabular}{ll|ccc|cc|cc}
\toprule[1.5pt]
\multicolumn{2}{c}{} &
\multicolumn{3}{c}{\textbf{Mathematics Reasoning\uparrow}} &
\multicolumn{2}{c}{\textbf{Visual Reasoning\uparrow}} &
\multicolumn{2}{c}{\textbf{Multimodal Composition\uparrow}} \\
\cmidrule(lr){3-5} \cmidrule(lr){6-7} \cmidrule(lr){8-9}
\textbf{Method} & \textbf{Model} &
MathVista$_{\text{mini}}$ &
MathVision$_{\text{mini}}$ &
MM-Math &
HallusionBench &
MMVP & MMStar &
ScienceQA \\
\midrule

Vanilla &  & 61.2 & 24.3 & 40.1 & 67.9 & 69.1 & 62.4 & 53.3 \\
Multimodal COT &  & 59.4 & 25.8 & 42.2 & 68.3 & 72.5 & 65.2 & 52.9  \\
CCOT &  & 64.4 & 25.1 & 41.7 & 68.5 & 71.8 & 64.7 & 54.1 \\
ICoT &  & 64.1 & 23.9 & 43.6 & 68.9 & 73.3 & 63.7 & 53.6\\
\rowcolor{lightblue}
\textbf{\textit{+DMLR (Ours)}} & 
\multirow[c]{-5}{*}[1pt]{\rotatebox{90}{\parbox{2.5cm}{\centering \fontsize{11pt}{13pt}\selectfont Qwen2.5 \\VL 7B}}}
 & \textbf{65.3} \up{4.1\%} & \textbf{26.7} \up{2.4\%} & \textbf{44.1} \up{4.0\%} & \textbf{70.4} \up{3.5\%} & \textbf{73.9} \up{4.8\%} & \textbf{67.1} \up{4.7\%} & \textbf{58.4} \up{5.1\%} \\
\midrule\midrule

Vanilla &  & 50.5 & 18.2 & 33.4 & 67.1 & 57.9 & 52.7 & 45.9 \\
Multimodal COT &  & 52.2 & 19.4 & 33.0 & 66.9 & 57.5 & 52.9 & 46.3 \\
CCOT & & 54.3 & 19.7 & 35.1 & 66.5 & 58.6 & 53.3 & 46.1 \\
ICoT &  & 54.0 & 19.3 & 34.6 & 67.3 & 58.4 & 53.0 & 46.7 \\
\rowcolor{lightblue}
\textbf{\textit{+DMLR (Ours)}} &
\multirow[c]{-5}{*}[1pt]{\rotatebox{90}{\parbox{2.5cm}{\centering \fontsize{11pt}{13pt}\selectfont Qwen2.5 \\VL 3B}}}
 & \textbf{55.3} \up{4.8\%} & \textbf{20.9} \up{2.7\%} & \textbf{36.3} \up{2.9\%} & \textbf{69.5} \up{2.4\%} & \textbf{59.2} \up{1.3\%} & \textbf{53.7} \up{1.00\%} & \textbf{48.1} \up{2.2\%}\\
\bottomrule[1.5pt]
\end{tabular}
}
\vspace{-0.5em}
\label{pass}
\end{table*}

\section{Theoretical Analyses}
\label{d}
\subsection{Confidence and Quality Gradient Alignment.}

\subsubsection{Problem Setup and Assumptions}
\noindent We consider the latent reasoning state $h \in \mathbb{R}^d$ and two scalar functions:
a confidence function $C : \mathbb{R}^d \to \mathbb{R}$ and a reasoning-quality function
$Q : \mathbb{R}^d \to \mathbb{R}$. During test-time optimization, DMLR updates the latent
state by ascending the confidence landscape:
\begin{equation}
    h^{+} = h + \eta \nabla C(h),
    \quad \eta > 0.
    \label{eq:dmlr_update_conf}
\end{equation}

\paragraph{Assumption A.1 (Differentiability).}
The functions $C$ and $Q$ are twice continuously differentiable on an open set 
$\mathcal{D} \subset \mathbb{R}^d$.
\vspace{-0.2cm}
\paragraph{Assumption A.2 (Smoothness of $Q$).}
There exists a constant $L_Q > 0$ such that
\begin{equation}
    \|\nabla Q(h_1) - \nabla Q(h_2)\|
    \;\le\;
    L_Q \,\|h_1 - h_2\|,
    \quad \forall\, h_1, h_2 \in \mathcal{D}.
    \label{eq:L_smooth_Q}
\end{equation}
\vspace{-0.8cm}
\paragraph{Assumption A.3 (Local Gradient Alignment).}
At a given point $h \in \mathcal{D}$, we have
\begin{equation}
    \nabla C(h)\cdot\nabla Q(h) > 0,
    \qquad
    \nabla C(h) \neq 0.
    \label{eq:local_alignment}
\end{equation}

\subsubsection{Auxiliary Inequality for $L_Q$--Smooth Functions}
\begin{lemma}[Descent lemma form]
\label{lem:descent_lemma}
Under Assumptions~A.1 and~A.2, for any $h \in \mathcal{D}$ and any vector 
$\Delta h$ such that $h + \Delta h \in \mathcal{D}$, we have
\begin{equation}
    Q(h + \Delta h)
    \;\ge\;
    Q(h) + \nabla Q(h)\cdot \Delta h
    - \frac{L_Q}{2}\,\|\Delta h\|^2.
    \label{eq:descent_lemma}
\end{equation}
\end{lemma}

\begin{proof}
Define a scalar function $g : [0,1] \to \mathbb{R}$ by
\begin{equation}
    g(t) = Q(h + t \Delta h).
\end{equation}
By Assumption~A.1, $g$ is twice continuously differentiable. By the chain rule,
\begin{equation}
    g'(t) = \nabla Q(h + t \Delta h)\cdot \Delta h.
\end{equation}
Using the fundamental theorem of calculus,
\begin{equation}
\begin{aligned}
    Q(h + \Delta h) - Q(h)
    &= g(1) - g(0)
     = \int_0^1 g'(t)\,\mathrm{d}t \\
    &= \int_0^1 \nabla Q(h + t \Delta h)\cdot\Delta h\,\mathrm{d}t.
\end{aligned}
\label{eq:Q_diff_integral}
\end{equation}
Add and subtract $\nabla Q(h)$ inside the integrand:
\begin{equation}
\small
\begin{aligned}
Q(h+\Delta h)-Q(h)
&= \int_{0}^{1}
\!\bigl[\nabla Q(h)
  + (\nabla Q(h+t\Delta h)-\nabla Q(h))\bigr]\!\cdot\Delta h\,\mathrm{d}t \\
&= \nabla Q(h)\!\cdot\!\Delta h
 + \int_{0}^{1}\!(\nabla Q(h+t\Delta h)-\nabla Q(h))\!\cdot\!\Delta h\,\mathrm{d}t .
\end{aligned}
\label{eq:Q_diff_split}
\end{equation}

By the Cauchy--Schwarz inequality and Assumption~A.2,
\begin{equation}
\small
\begin{aligned}
\bigl|(\nabla Q(h+t\Delta h)-\nabla Q(h))\!\cdot\!\Delta h\bigr|
&\le \|\nabla Q(h+t\Delta h)-\nabla Q(h)\|\,\|\Delta h\| \\
&\le L_Q\, t\, \|\Delta h\|^2 .
\end{aligned}
\end{equation}

Hence,
\begin{equation}
\small
\int_{0}^{1} (\nabla Q(h+t\Delta h)-\nabla Q(h))\!\cdot\!\Delta h\,dt
\;\ge\;
- \int_{0}^{1} L_Q t\,\|\Delta h\|^{2}\,dt
= -\tfrac{L_Q}{2}\|\Delta h\|^{2}.
\end{equation}

\noindent Substituting into~\eqref{eq:Q_diff_split} yields~\eqref{eq:descent_lemma}, which completes the proof. \qedhere
\end{proof}

\subsubsection{Proof of Theorem 4.1}

We now prove the local, step-size--bounded positive correlation between confidence and reasoning quality.

\begin{theorem}[Local bounded positive gain]
\label{thm:local_positive_correlation}
Let Assumptions~A.1--A.3 hold at $h \in \mathcal{D}$. Define
\begin{equation}
    a := \nabla C(h)\cdot\nabla Q(h) > 0,
    \qquad
    b := \|\nabla C(h)\|^2 > 0.
    \label{eq:def_a_b}
\end{equation}
Then for any step size $\eta$ satisfying
\begin{equation}
    0 < \eta < \frac{2a}{L_Q b},
    \label{eq:eta_bound}
\end{equation}
the update~\eqref{eq:dmlr_update_conf} satisfies
\begin{equation}
    Q(h^{+}) > Q(h).
    \label{eq:Q_increase}
\end{equation}
In other words, within the bounded step-size interval~\eqref{eq:eta_bound}, a single step of confidence ascent strictly improves the reasoning-quality function $Q$.
\end{theorem}

\begin{proof}
Apply Lemma~\ref{lem:descent_lemma} with $\Delta h = \eta \nabla C(h)$, which gives
\begin{equation}
\begin{aligned}
    Q(h^{+}) - Q(h)
    &\ge 
    \nabla Q(h)\cdot (\eta \nabla C(h))
    - \frac{L_Q}{2}\eta^2 \|\nabla C(h)\|^2 \\
    &= \eta \,\nabla Q(h)\cdot\nabla C(h)
    - \frac{L_Q}{2}\eta^2 \|\nabla C(h)\|^2.
\end{aligned}
\label{eq:Q_diff_eta}
\end{equation}
Using the definitions in~\eqref{eq:def_a_b}, we obtain
\begin{equation}
    Q(h^{+}) - Q(h)
    \;\ge\;
    \eta a - \frac{L_Q}{2}\eta^2 b
    \;=:\;
    \phi(\eta).
    \label{eq:phi_eta}
\end{equation}
The function $\phi(\eta)$ is a concave quadratic in $\eta$ with roots at $\eta=0$ and
\begin{equation}
    \eta^\star = \frac{2a}{L_Q b}.
\end{equation}
Since $a>0$ and $b>0$, we have $\eta^\star>0$ and
\begin{equation}
    \phi(\eta) > 0
    \quad\text{for all } 0<\eta<\eta^\star.
\end{equation}
Therefore, for any $0<\eta<\eta^\star$, we obtain
\begin{equation}
    Q(h^{+}) - Q(h) \ge \phi(\eta) > 0,
\end{equation}
which proves~\eqref{eq:Q_increase}. 
\end{proof}

\subsubsection{Boundedness of the Positive-Correlation Region}

Theorem~\ref{thm:local_positive_correlation} shows that, at any point where the gradients of 
$C$ and $Q$ are positively aligned, there exists a \emph{bounded} step-size interval within which
ascending the confidence landscape strictly improves $Q$. We next argue that such a
positive-correlation property cannot hold globally in the presence of a ``confidently incorrect'' trap.

\begin{definition}[Confidently incorrect trap]
\label{def:trap}
A point $h_{\mathrm{trap}} \in \mathcal{D}$ is called a confidently incorrect trap if
\begin{align}
    &\nabla C(h_{\mathrm{trap}}) = 0,
    \quad
    \nabla^2 C(h_{\mathrm{trap}}) 
    \text{ is negative definite,}
    \label{eq:trap_conf_local_max} \\
    &Q(h_{\mathrm{trap}}) \ll \max_{h\in\mathcal{D}} Q(h).
    \label{eq:trap_low_quality}
\end{align}
In other words, $h_{\mathrm{trap}}$ is a strict local maximizer of the confidence landscape, but the corresponding reasoning quality is significantly lower than that of other regions.
\end{definition}

Consider the dynamical system induced by confidence ascent,
\begin{equation}
    h^{(t+1)} = h^{(t)} + \eta \nabla C\big(h^{(t)}\big),
    \label{eq:conf_dynamics}
\end{equation}
with a sufficiently small fixed step size $\eta>0$. Let $\mathcal{B}(h_{\mathrm{trap}})$
denote the basin of attraction of $h_{\mathrm{trap}}$ under~\eqref{eq:conf_dynamics}, i.e.,
the set of initial points $h^{(0)}$ for which $h^{(t)} \to h_{\mathrm{trap}}$ as $t\to\infty$.

\begin{proposition}[Positive correlation cannot hold everywhere in the basin]
\label{prop:bounded_region}
Suppose there exists a confidently incorrect trap $h_{\mathrm{trap}}$ in the sense of
Definition~\ref{def:trap}. Then it is impossible that
\begin{equation}
    \nabla C(h)\cdot\nabla Q(h) > 0,
    \quad \forall\, h \in \mathcal{B}(h_{\mathrm{trap}}).
    \label{eq:global_alignment_basin}
\end{equation}
Equivalently, there must exist points $h$ in the basin of attraction for which
\begin{equation}
    \nabla C(h)\cdot\nabla Q(h) \le 0.
\end{equation}
Thus, the region where confidence and reasoning-quality gradients are positively aligned
is necessarily bounded and cannot cover the entire basin of a confidently incorrect trap.
\end{proposition}

\begin{proof}
Assume, for the sake of contradiction, that~\eqref{eq:global_alignment_basin} holds, i.e.,
for all $h \in \mathcal{B}(h_{\mathrm{trap}})$ we have $\nabla C(h)\cdot\nabla Q(h) > 0$.
Pick an arbitrary $h^{(0)} \in \mathcal{B}(h_{\mathrm{trap}})$, and consider the trajectory
generated by~\eqref{eq:conf_dynamics}:
\begin{equation}
    h^{(t+1)} = h^{(t)} + \eta \nabla C(h^{(t)}).
\end{equation}

\noindent By Definition of $\mathcal{B}(h_{\mathrm{trap}})$, we have
\begin{equation}
    h^{(t)} \to h_{\mathrm{trap}}
    \quad\text{as } t\to\infty.
\end{equation}
Since $Q$ is continuous, it follows that
\begin{equation}
    \lim_{t\to\infty} Q(h^{(t)}) = Q(h_{\mathrm{trap}}).
    \label{eq:limit_Q_trap}
\end{equation}

\noindent On the other hand, for each $t$ such that $h^{(t)}\in\mathcal{D}$, Assumptions~A.1--A.2
hold at $h^{(t)}$ and the alignment condition
$\nabla C(h^{(t)})\cdot\nabla Q(h^{(t)})>0$ holds by~\eqref{eq:global_alignment_basin}. 
Hence, by Theorem~\ref{thm:local_positive_correlation}, there exists a step-size bound
$\eta_t^\star>0$ such that for all $0<\eta<\eta_t^\star$,
\begin{equation}
    Q(h^{(t+1)}) > Q(h^{(t)}).
\end{equation}
Taking $\eta$ small enough so that $\eta < \eta_t^\star$ for all $t$ along the trajectory,
we obtain a strictly increasing sequence:
\begin{equation}
    Q(h^{(0)}) < Q(h^{(1)}) < Q(h^{(2)}) < \cdots.
\end{equation}

Thus $\{Q(h^{(t)})\}_{t=0}^\infty$ is strictly increasing, and together with
\eqref{eq:limit_Q_trap}, we obtain $Q(h_{\mathrm{trap}})>Q(h^{(0)})$. 
This contradicts the low-quality condition \eqref{eq:trap_low_quality}, since
$h_{\mathrm{trap}}$ cannot be a confidently incorrect trap. 
Therefore, \eqref{eq:global_alignment_basin} must be false, and there exist
points $h\in\mathcal{B}(h_{\mathrm{trap}})$ such that
$\nabla C(h)\cdot\nabla Q(h)\le 0$. 
This completes the proof.
\end{proof}

\noindent\textbf{\textit{Proposition~\ref{prop:bounded_region} shows that the positive correlation between confidence
and reasoning quality, formalized via gradient alignment, is fundamentally a \emph{local}
property: it can hold within a bounded region (and within bounded step sizes) but cannot be
extended to the entire latent space in the presence of confidently incorrect traps.}}

\subsection{Visual Injection Enhances Confidence.}

Let $\mathcal{T}$ denote the latent reasoning states before visual injection,
$\hat{\mathcal{T}}$ the updated states after dynamic visual injection (DVI), and
$z_v$ the visual features.  
Let $Y$ be the target output random variable (e.g., the final answer).
We assume all variables are defined on a common probability space.

\noindent \textbf{Assumption B.1.}
The dynamic visual injection (DVI) module is constructed such that the
latent state after injection preserves at least as much visual information
as before:
\begin{equation}
    I(\hat{\mathcal{T}}; z_v) \;\ge\; I(\mathcal{T}; z_v).
    \label{eq:B1}
\end{equation}

\noindent \textbf{Assumption B.2.}
The visual features contain information about the target variable $Y$:
\begin{equation}
    I(z_v; Y) > 0.
    \label{eq:B2_visionY}
\end{equation}
Moreover, the task representation is trained so that the mutual information
between the latent state and $Y$ is a strictly increasing function of the
mutual information between the latent state and visual features.  
That is, there exists a strictly increasing function $g(\cdot)$ satisfying
\begin{equation}
    I(Y; \mathcal{T}) = g\!\big(I(\mathcal{T}; z_v)\big),\qquad
    I(Y; \hat{\mathcal{T}}) = g\!\big(I(\hat{\mathcal{T}}; z_v)\big).
    \label{eq:B2_g_function}
\end{equation}

\noindent \textbf{Assumption B.3.}
The model’s confidence objective is a strictly decreasing function of the
conditional entropy of $Y$ given the latent state:
\begin{equation}
    J_{\mathrm{conf}}(\mathcal{T}) 
    = \psi\!\big(H(Y\mid \mathcal{T})\big),\qquad
    J_{\mathrm{conf}}(\hat{\mathcal{T}})
    = \psi\!\big(H(Y\mid \hat{\mathcal{T}})\big),
    \label{eq:B3_psi}
\end{equation}
where $\psi(\cdot)$ is strictly decreasing.  
This reflects that lower predictive uncertainty corresponds to higher confidence.

\noindent We now show that~\eqref{eq:B1} implies
$J_{\mathrm{conf}}(\hat{\mathcal{T}})\ge J_{\mathrm{conf}}(\mathcal{T})$.

\begin{lemma}[From vision MI to task MI]
\label{lem:MI_to_task}
Under Assumption~B.2, if
$I(\hat{\mathcal{T}}; z_v)\ge I(\mathcal{T}; z_v)$, then
\begin{equation}
    I(Y;\hat{\mathcal{T}})\;\ge\; I(Y;\mathcal{T}).
    \label{eq:task_MI_increase}
\end{equation}
\end{lemma}

\begin{proof}
By~\eqref{eq:B2_g_function},
\[
I(Y;\hat{\mathcal{T}}) = g\!\big(I(\hat{\mathcal{T}}; z_v)\big),
\quad
I(Y;\mathcal{T}) = g\!\big(I(\mathcal{T}; z_v)\big).
\]
Since $g$ is strictly increasing and
$I(\hat{\mathcal{T}}; z_v)\ge I(\mathcal{T}; z_v)$, we obtain~\eqref{eq:task_MI_increase}.
\end{proof}

\begin{lemma}[Task MI and conditional entropy]
\label{lem:MI_entropy}
If $I(Y;\hat{\mathcal{T}})\ge I(Y;\mathcal{T})$, then
\begin{equation}
    H(Y\mid \hat{\mathcal{T}}) \;\le\; H(Y\mid \mathcal{T}).
    \label{eq:entropy_decrease}
\end{equation}
\end{lemma}

\begin{proof}
By the identity $I(Y;U) = H(Y) - H(Y\mid U)$ for any random variable $U$,
\[
H(Y\mid U) = H(Y) - I(Y;U).
\]
Thus
\[
H(Y\mid \hat{\mathcal{T}}) = H(Y) - I(Y;\hat{\mathcal{T}}),
\quad
H(Y\mid \mathcal{T}) = H(Y) - I(Y;\mathcal{T}).
\]
If $I(Y;\hat{\mathcal{T}})\ge I(Y;\mathcal{T})$, then
$H(Y\mid \hat{\mathcal{T}})\le H(Y\mid \mathcal{T})$, which is~\eqref{eq:entropy_decrease}.
\end{proof}

\begin{lemma}[Conditional entropy and confidence]
\label{lem:entropy_confidence}
Under Assumption~B.3, if 
$H(Y\mid \hat{\mathcal{T}})\le H(Y\mid \mathcal{T})$, then
\begin{equation}
    J_{\mathrm{conf}}(\hat{\mathcal{T}})
    \;\ge\;
    J_{\mathrm{conf}}(\mathcal{T}).
    \label{eq:conf_increase}
\end{equation}
\end{lemma}

\begin{proof}
By~\eqref{eq:B3_psi},
\[
J_{\mathrm{conf}}(\hat{\mathcal{T}})
= \psi\!\big(H(Y\mid \hat{\mathcal{T}})\big),
\quad
J_{\mathrm{conf}}(\mathcal{T})
= \psi\!\big(H(Y\mid \mathcal{T})\big).
\]
Since $\psi$ is strictly decreasing and
$H(Y\mid \hat{\mathcal{T}})\le H(Y\mid \mathcal{T})$,
we obtain~\eqref{eq:conf_increase}.
\end{proof}

\begin{theorem}[For Theorem 4.2]
\label{thm:theorem42_formal}
Under Assumptions~B.1--B.3, if
\begin{equation}
    I(\hat{\mathcal{T}}; z_v)\;\ge\; I(\mathcal{T}; z_v),
\end{equation}
then
\begin{equation}
    J_{\mathrm{conf}}(\hat{\mathcal{T}})
    \;\ge\;
    J_{\mathrm{conf}}(\mathcal{T}).
\end{equation}
\end{theorem}

\begin{proof}
From Assumption~B.1 and Lemma~\ref{lem:MI_to_task},
$I(\hat{\mathcal{T}}; z_v)\ge I(\mathcal{T}; z_v)$
implies $I(Y;\hat{\mathcal{T}})\ge I(Y;\mathcal{T})$.
By Lemma~\ref{lem:MI_entropy}, this yields
$H(Y\mid \hat{\mathcal{T}})\le H(Y\mid \mathcal{T})$.
Finally, Lemma~\ref{lem:entropy_confidence} implies
$J_{\mathrm{conf}}(\hat{\mathcal{T}})\ge J_{\mathrm{conf}}(\mathcal{T})$.
\end{proof}

\noindent Theorem~\ref{thm:theorem42_formal} formalizes that, under mild assumptions,
DVI-induced increases in the mutual information between latent states and visual
features translate into lower predictive uncertainty about the target output,
and thus into higher expected confidence.

\subsection{Derivation of the Latent Policy Gradient and Update Rule}
\label{app:latent_pg_full}
We give a detailed derivation of the gradient used to update the latent
thought vectors $H$ (e.g., latent think tokens) via a Gaussian policy gradient method.
\vspace{-0.2cm}
\paragraph{Objective.}
Let $A$ denote the sampled latent action (a perturbed latent thought
vector) and $R(A)$ the corresponding reward or confidence score. Given a
stochastic policy $\pi(A\mid H)$ over actions $A$ conditioned on the
current latent vector $H$, we define the expected objective
\begin{equation}
    J(H)
    := \mathbb{E}_{A\sim\pi(\cdot\mid H)}[R(A)].
    \label{eq:latent_obj}
\end{equation}

Our goal is to derive $\nabla_H J(H)$ and obtain a practical estimator
for use in gradient ascent.

\subsubsection{Policy Gradient Form}
\begin{lemma}[Policy gradient identity]
\label{lem:pg_identity_H}
Under mild regularity assumptions (allowing differentiation under the
integral sign), the gradient of $J(H)$ in~\eqref{eq:latent_obj} satisfies
\begin{equation}
    \nabla_H J(H)
    = \mathbb{E}_{A\sim\pi(\cdot\mid H)}
      \big[ R(A)\,\nabla_H \log \pi(A\mid H) \big].
    \label{eq:pg_theorem_H}
\end{equation}
\end{lemma}

\begin{proof}
By definition of expectation,
\begin{align}
    J(H)
    &= \int R(A)\,\pi(A\mid H)\,dA.
\end{align}
Differentiating w.r.t.\ $H$ yields
\begin{align}
    \nabla_H J(H)
    &= \int R(A)\,\nabla_H \pi(A\mid H)\,dA.
\end{align}
Using the log-derivative identity
$\nabla_H \pi(A\mid H) = \pi(A\mid H)\,\nabla_H \log\pi(A\mid H)$, we obtain
\begin{align}
    \nabla_H J(H)
    &= \int R(A)\,\pi(A\mid H)\,\nabla_H \log\pi(A\mid H)\,dA \\
    &= \mathbb{E}_{A\sim\pi(\cdot\mid H)}
       \big[ R(A)\,\nabla_H \log \pi(A\mid H) \big],
\end{align}
which is~\eqref{eq:pg_theorem_H}.
\end{proof}

\subsubsection{Gaussian Latent Policy}
We now instantiate the policy as an isotropic Gaussian perturbation of
$H$. Specifically, we sample
\begin{equation}
    A = H + \xi^{(t)},
    \qquad
    \xi^{(t)} \sim \mathcal{N}(0,\sigma^{2}I),
    \label{eq:gaussian_policy_A}
\end{equation}
which is equivalent to
\begin{equation}
    \pi(A\mid H)
    = \mathcal{N}(A;\,H,\sigma^{2}I).
\end{equation}

\noindent We next compute the gradient of the log-probability under this Gaussian
policy.

\begin{lemma}[Gradient of Gaussian log-probability]
\label{lem:gaussian_score}
For the Gaussian policy in~\eqref{eq:gaussian_policy_A}, the
log-probability gradient satisfies
\begin{equation}
    \nabla_H \log \pi(A\mid H)
    = \frac{A - H}{\sigma^{2}}
    = \frac{\xi}{\sigma^{2}}.
    \label{eq:gaussian_score}
\end{equation}
\end{lemma}

\begin{proof}
The log-density of $\pi(A\mid H)$ is
\begin{equation}
    \log\pi(A\mid H)
    = -\frac{1}{2\sigma^{2}}\|A-H\|^{2} + \mathrm{const},
\end{equation}
where the constant does not depend on $H$. Differentiating w.r.t.\ $H$,
\begin{align}
    \nabla_H \log\pi(A\mid H)
    &= -\frac{1}{2\sigma^{2}}
       \nabla_H \|A-H\|^{2} \\
    &= -\frac{1}{2\sigma^{2}}
       \nabla_H (A-H)^\top(A-H) \\
    &= -\frac{1}{2\sigma^{2}}\cdot(-2)(A-H) \\
    &= \frac{A-H}{\sigma^{2}}.
\end{align}
Using $A = H + \xi$ from~\eqref{eq:gaussian_policy_A} yields
$A-H = \xi$, giving~\eqref{eq:gaussian_score}.
\end{proof}

\noindent  Substituting~\eqref{eq:gaussian_score} into the policy gradient
expression~\eqref{eq:pg_theorem_H}, we obtain
\begin{align}
    \nabla_H J(H)
    &= \mathbb{E}_{A\sim\pi(\cdot\mid H)}
       \big[ R(A)\,\nabla_H \log\pi(A\mid H) \big] \\
    &= \mathbb{E}_{\epsilon\sim\mathcal{N}(0,\sigma^{2}I)}
       \left[
           R(H+\xi)\,\frac{\xi}{\sigma^{2}}
       \right].
    \label{eq:latent_pg_gaussian}
\end{align}

\subsubsection{Monte Carlo Estimator from a Single Sample}

In practice, the expectation in~\eqref{eq:latent_pg_gaussian} is
approximated via Monte Carlo sampling. At optimization step $t$, we draw
a single Gaussian noise sample
\[
\mathcal{\xi}^{(t)}\sim \mathcal{N}(0,\sigma^{2}I),
\qquad
A^{(t)} = H^{(t)} + \xi^{(t)},
\]
and construct the one-sample estimator
\begin{equation}
    \nabla_H J(H^{(t)})
    \approx R\big(H^{(t)} + \xi^{(t)}\big)
             \frac{\xi^{(t)}}{\sigma^{2}}.
    \label{eq:latent_pg_MC}
\end{equation}

\begin{lemma}[Unbiasedness of the Monte Carlo estimator]
\label{lem:unbiased_MC}
Let $\mathcal{\xi}^{(t)}\sim \mathcal{N}(0,\sigma^{2}I)$ and define the
estimator
\(
    \widehat{g}^{(t)}
    = R(H^{(t)} + \xi^{(t)})
      \xi^{(t)}/\sigma^{2}.
\)
Then
\begin{equation}
    \mathbb{E}[\widehat{g}^{(t)}]
    = \nabla_H J(H^{(t)}),
\end{equation}
where the expectation is taken over $\xi^{(t)}$.
\end{lemma}

\begin{proof}
By definition and linearity of expectation,
\begin{align}
    \mathbb{E}[\widehat{g}^{(t)}]
    &= \mathbb{E}_{\xi^{(t)}}
       \left[
           R(H^{(t)} + \xi^{(t)})
           \frac{\epsilon^{(t)}}{\sigma^{2}}
       \right]
     = \nabla_H J(H^{(t)}),
\end{align}
where the last equality is exactly
\eqref{eq:latent_pg_gaussian} evaluated at $H = H^{(t)}$.
\end{proof}

\subsubsection{Gradient Ascent Update}

Finally, we perform standard gradient ascent on the latent vector $H$.
With learning rate $\eta>0$, the update at step $t$ is
\begin{equation}
    H^{(t+1)}
    = H^{(t)}
      + \eta \cdot
        R\big(H^{(t)} + \xi^{(t)}\big)
        \frac{\xi^{(t)}}{\sigma^{2}}.
    \label{eq:latent_update_rule}
\end{equation}

\noindent This update moves the latent thought vector $H$ in the direction of
noisy perturbations $\xi^{(t)}$ that lead to higher reward, as
estimated by the Monte Carlo policy gradient.

\section{More detail about Visual Injection.}
\label{e}

\textbf{Different Strategies Discussion.} We explore two visual injection mechanisms to identify which strategy best supports multi-step latent reasoning optimization. \textbf{\textit{The first strategy}}, ultimately adopted in our DVI Strategy is a dynamic visual patch search-and-refinement procedure organized around the three-phase cycle shown in Algorithm~\ref{latent_reasoning}. At the start of the reasoning trajectory, we use the initial latent think token’s cross-modal attention to select the top-$m$ visually relevant patches, forming an initial candidate set $\mathcal{Z}_{\text{cand}}$ and an initial best-patch set $\mathcal{V}_{\text{best}}$. This initialization provides a reasonable but unconverged visual hypothesis, ensuring that the search begins from semantically meaningful regions while preserving room for subsequent refinement.

During each latent-update iteration, we recompute attention based on the updated hidden state $\mathcal{T}^{(t)}$ and resample a fresh batch of candidate patches $\mathcal{Z}_{\text{cand}} = \{Z_1, \dots, Z_m\}$. This dynamic resampling allows the visual hypothesis to adapt in tandem with the evolving reasoning context, preventing the model from becoming locked to the potentially imperfect initial attention pattern. We then construct an expanded latent representation $\tilde{\mathcal{T}}^{(t)}_l = [\mathcal{T}^{(t)}_l, \mathcal{Z}_{\text{cand}}, \mathcal{V}_{\text{best}}]$ and compute its reward $r = \mathcal{R}(\mathcal{Z}, \mathcal{Q}, \tilde{\mathcal{T}}^{(t)}_l)$, which acts as a task-level criterion for validating whether the newly sampled patches truly enhance reasoning performance.

If the obtained reward surpasses the historical best value $r_{\text{best}}$, we incorporate the new patches into the best-patch set $\mathcal{V}_{\text{best}}$ and carry forward the updated latent representation into the next iteration; otherwise, we revert the update to preserve previously verified visual evidence. Over multiple iterations, $\mathcal{V}_{\text{best}}$ is progressively refined under the interplay of exploration (resampling new patches) and exploitation (reward-based filtering), ultimately converging to a compact and semantically stable set of visual anchors. This dynamic best-patch optimization provides consistent and reward-validated visual grounding throughout the reasoning chain, producing far more reliable behavior than strategies that rely solely on per-step attention signals.

\textbf{\textit{The second strategy}} adopts a more direct formulation: at each iteration, it simply injects the visual patch that receives the highest attention, thereby emulating the attention-guided visual grounding commonly used in existing multimodal models. These two strategies thus represent fundamentally different paradigms for visual injection—one progressively optimizes visual evidence through reward-driven refinement, whereas the other relies on instantaneous attention-based selection. \textit{Given their distinct underlying principles, it is necessary to systematically compare their behaviors and performance characteristics.}

\noindent \textbf{Results Analysis.} As shwon in Table~\ref{tab:stability_compare} and \ref{tab:perf_compare}, we observe that injecting the highest attention visual patch at each iteration appears straightforward but suffers from fundamental instability. Attention scores are highly local and easily influenced by visual saliency, token co-adaptation, and early stage noise, causing the most attended region to deviate from the truly task relevant area. Because this strategy depends entirely on instantaneous attention distributions, the injected visual evidence fluctuates across iterations, resulting in drifting visual focus and compounding errors when early attention is misaligned. In contrast, our dynamic visual injection strategy, based on iterative patch resampling, reward evaluation, and best patch refinement, provides a far more stable grounding mechanism. By validating new patches through task level reward rather than noisy attention signals, the model progressively converges to a semantically consistent and task relevant set of visual anchors. This search, verify, refine process stabilizes visual grounding across iterations, prevents error amplification, and yields markedly more reliable reasoning trajectories. Empirically, the dynamic best patch optimization consistently outperforms attention based injection, demonstrating superior robustness and significantly higher multimodal reasoning accuracy.

\begin{table}[h]
\centering
\caption{Performance comparison between the two visual injection strategies on the MMVP Benchmark.}
\vspace{0.1cm}
\resizebox{0.8\linewidth}{!}{
\begin{tabular}{l c}
\toprule
\textbf{Method} & \textbf{Accuracy $\uparrow$} \\
\midrule
Attention-based  & 66.3 \\
Dynamic best-patch (ours) & \textbf{74.8} \\
\bottomrule
\end{tabular}}
\label{tab:perf_compare}
\end{table}

\begin{table}[h]
\centering
\caption{Stability metrics across iterative reasoning on the MathVision Benchmark. Lower is better.}
\vspace{0.1cm}
\resizebox{\linewidth}{!}{
\begin{tabular}{l cc}
\toprule
\textbf{Method} & \textbf{Attention Variance $\downarrow$} & \textbf{Drift Distance $\downarrow$} \\
\midrule
Attention-based  & 0.48 & 12.3 \\
Dynamic best-patch (ours) & \textbf{0.12} & \textbf{2.1} \\
\bottomrule
\end{tabular}}
\label{tab:stability_compare}
\end{table}

\noindent \textbf{In-depth Analysis.} When designing DVI, a seemingly reasonable question arises: since visual injection is intended to supply additional evidence when the model becomes uncertain, can confidence be used as a trigger signal so that visual patch optimization is performed only when confidence falls below a predefined threshold? This confidence-triggered variant appears more economical and aligns with the intuition of “inject vision only when needed.” However, our experiments (as shown in Table~\ref{tab:confidence_vs_continuous}) reveal that this strategy leads to noticeably worse convergence behavior. The optimization of visual patches in DVI is inherently continuous and incremental, relying on multi-step accumulation rather than sporadic corrections. Restricting updates to low confidence moments breaks this continuity. In practice, the model often enters a suboptimal regime in which confidence remains superficially stable while the underlying visual grounding is still misaligned, delaying or even preventing necessary corrections. As a result, the overall number of iterations required for convergence increases, and the final visual anchor becomes less stable and less semantically consistent than under the continuous best-patch refinement strategy. These findings indicate that confidence is not a reliable indicator for when visual refinement should occur, and maintaining per-iteration optimization is essential for achieving robust, efficient, and semantically coherent convergence in DVI.

\begin{table}[t]
\centering
\caption{
Comparison between confidence-triggered refinement and continuous DVI optimization on the MMVP Benchmark.
\textbf{Iter.} refers to the number of iterations required for convergence,
\textbf{Cons.} refers to the cross-iteration patch consistency,
\textbf{Err.} refers to the grounding error,
and \textbf{Acc.} refers to the final answer accuracy.
Continuous optimization converges faster, produces more stable visual anchors, and achieves higher accuracy.
}
\vspace{0.15cm}
\resizebox{\linewidth}{!}{
\begin{tabular}{lcccc}
\toprule
\textbf{Method} &
\textbf{Iter. $\downarrow$} &
\textbf{Cons. $\uparrow$} &
\textbf{Err. $\downarrow$} &
\textbf{Acc. $\uparrow$} \\
\midrule
Confidence-triggered update & 14.2 & 0.46 & 0.27 & 68.1 \\
Continuous best-patch (ours) & \textbf{9.3} & \textbf{0.83} & \textbf{0.12} & \textbf{74.8} \\
\bottomrule
\end{tabular}}
\label{tab:confidence_vs_continuous}
\vspace{-0.2cm}
\end{table}

\begin{algorithm}[h] 
\caption{\small \textbf{\textit{Dynamic Multimodal Latent Reasoning}}}
\small
\label{latent_reasoning}
\DontPrintSemicolon
\SetAlgoVlined
\SetKwProg{Fn}{}{}{}
\KwRequire{Image embeddings $\mathcal{Z}$, text embeddings $\mathcal{Q}$, 
latent tokens $\mathcal{T}_l$, learning rate $\eta$, 
iterations $T$, best visual patch $\mathcal{V}_{\textit{best}}$, 
top-$k$ probability $\mathrm{Top}_k(\mathcal{P}_i)$, the number of candidate patches 
$m$}

\vspace{2pt}
$\mathrm{Top}_k(\mathcal{P}_i) = \pi_{\theta}([\mathcal{Q}, \mathcal{Z}, \mathcal{T}])$; 
$r \gets \mathcal{R}(\mathcal{P}_{i})$ 
\hspace{1.em}\textcolor{blue}{$\triangleright$ reward}

\vspace{3pt}
\textcolor{darkgray}{\# \textit{Latent Policy Gradient Optimization}}\;

\For{$T = 1 \dots t$}{
    $\epsilon \sim \mathcal{N}(0,\sigma^2 I)$ 
    \hfill \textcolor{blue}{$\triangleright$ latent perturbation}

    $\mathcal{T}^{(t)\prime} \gets \mathcal{T}^{(t)} + \epsilon$

    $\mathcal{T}^{(t)} \gets \mathcal{T}^{(t)} + 
    \eta \nabla_{\mathcal{T}^{(t)}} \mathcal{J}(\mathcal{T}^{(t)})$
    \hfill \textcolor{blue}{$\triangleright$ latent update}

    \vspace{2pt}
    \textcolor{darkgray}{\# \textit{Dynamic Visual Injection}}\;
    
    $\mathcal{V}_{\textit{best}} \gets \text{Initialize}(\mathcal{T}^{(0)}, m)$ \hfill \textcolor{blue}{$\triangleright$ initialize best patch}

    \For{$L = 1 \dots l$}{
    
       $\mathcal{Z}_{\textit{cand}} \gets 
        \text{AttentionSelect}(\mathcal{T}_l^{(t)}, m)$
        \hfill \textcolor{blue}{$\triangleright$ select $m$ candidate visual patches}

        $\tilde{\mathcal{T}_l}^{(t)} \gets [\mathcal{T}_l^{(t)}, \mathcal{Z}_{\textit{cand}}, \mathcal{V}_{\textit{best}}]$

        $r \gets \mathcal{R}(\mathcal{Q}, \mathcal{Z}, \tilde{\mathcal{T}^{(t)}_l})$

        \If{$r > r_{\text{best}}$}{
            $\mathcal{V}_{\textit{best}} \gets \mathcal{V}_{\textit{best}} \cup \mathcal{Z}_{\textit{cand}}$;

            $\mathcal{T}_l^{(t)} \gets \tilde{\mathcal{T}_l}^{(t)}$
            \hfill \textcolor{blue}{$\triangleright$ update best}
        }
        \Else{
            $\mathcal{T}_l^{(t)} \gets 
            [\mathcal{T}_l^{(t)}, \mathcal{V}_{\textit{best}}]$
            \hfill \textcolor{blue}{$\triangleright$ revert to previous best}
        }
        \vspace{1pt}
    }
    \vspace{2pt}
}

\vspace{3pt}
$\mathcal{X} \gets \text{Decode}(\mathcal{T}^{(t)}, \mathcal{Z}, \mathcal{Q})$

\Return $\mathcal{X}$
\end{algorithm}

\section{Case Study}
\label{f}
In this section, we provide qualitative examples to illustrate the behavior of the DMLR-enhanced model across different task settings. As shown in \textbf{Example 1-8}, the model demonstrates more reliable and fine-grained visual perception, supported by stable and semantically coherent internal representations. In reasoning scenarios, the model produces accurate and well-structured logical steps. Importantly, the dynamic latent reasoning mechanism suppresses unnecessary expansions, resulting in concise and task-focused reasoning chains rather than overly long or speculative explanations. Taken together, these examples highlight that DMLR simultaneously strengthens perceptual quality and reasoning reliability, while ensuring that the generated reasoning remains efficient and free from redundant elaboration.

\begin{figure*}[t]
    \centering
    \begin{mdframed}[
        frametitle={Perception Example 1},
        frametitlebackgroundcolor=lightyellow,
        backgroundcolor=gray!10,
        linecolor=gray,
        frametitlerule=true,
        frametitlefont=\color{black}\bfseries
    ]
        \centering
        \small
        \includegraphics[width=0.4\linewidth]{img/example/mmvp1.jpg} 
        \captionof{figure}{\centering Question: Is the door of the truck cab open?\\Options: (a) Yes (b) No\\Ground Truth: (a)       \label{fig1}} 
  
        \flushleft

    \noindent\textbf{Qwen2.5-VL-7B-Instruct + DMLR:}

    The image shows a garbage truck with its door on the side of the cab open. This is evident from the visible interior and the fact that the door is not aligned with the rest of the vehicle's structure, indicating it has been opened. The answer is \boxed{a}. 

    \vspace{1em}

    \noindent\textbf{Qwen2.5-VL-7B-Instruct:}

    To determine if the door of the truck cab is open, let's analyze the image step by step:
    \begin{enumerate}
        \item Observe the side of the truck where the door would be located.
        \item Look for any visible gap or opening that would indicate the door is not closed.
        \item Check the overall appearance of the door; if it is open, there should be a noticeable gap or difference in the alignment compared to when it is closed.
    \end{enumerate}

    Upon examining the image:
    \begin{itemize}
        \item The door appears to be aligned with the rest of the truck's body.
        \item There is no visible gap or difference in the alignment that would suggest the door is open.
    \end{itemize}

    Based on this analysis, the door of the truck cab does not appear to be open.

    \noindent\texttt{<think>} The door of the truck cab is not open.\texttt{</think>}
    \noindent\texttt{<answer>} No\texttt{</answer>}

    \noindent\boxed{\text{No}}
       \label{fig1}
    \end{mdframed}
\end{figure*}

\begin{figure*}[t]
    \centering
    \begin{mdframed}[
        frametitle={Perception Example 2},
        frametitlebackgroundcolor=lightyellow,
        backgroundcolor=gray!10,
        linecolor=gray,
        frametitlerule=true,
        frametitlefont=\color{black}\bfseries
    ]
        \centering
        \small
        \includegraphics[width=0.4\linewidth]{img/example/mmvp2.jpg} 
        \captionof{figure}{\centering Question: What is the woman more likely to do?\\Options: (a) Bump (b) Set\\Ground Truth: (b)} 
        \flushleft

    \noindent\textbf{Qwen2.5-VL-7B-Instruct + DMLR:}

    In the image, a volleyball game is in progress. The woman in focus appears to be positioned near the net, likely preparing to receive the ball. In volleyball, there are two primary actions for receiving the ball: bumping and setting.

    \begin{itemize}
        \item Bumping involves using the hands to cushion the ball, typically when it's coming down from above.
        \item Setting involves using the hands to lift the ball upward, usually when it's coming down or is at a low trajectory.
    \end{itemize}

    Given her position and the context of the game, she seems to be preparing to set the ball rather than bump it. This is because setting is a more strategic action that allows the team to control the direction and height of the ball, which is crucial for passing and spiking effectively.

    Therefore, based on the context and typical volleyball gameplay, the woman is more likely to set the ball. The answer is \boxed{b) \text{Set}}.

    \vspace{1em}

    \noindent\textbf{Qwen2.5-VL-7B-Instruct:}

    To determine what the woman is more likely to do, let's analyze the image step by step:
    \begin{enumerate}
        \item \textbf{Identify the context}: The image shows a volleyball game in progress. The woman in the foreground appears to be preparing to receive the ball, which is above her head.
        \item \textbf{Understand the volleyball terms}:
        \begin{itemize}
            \item \textbf{Bumping}: This is when a player uses their hands to pass the ball back into play after it has been spiked or set.
            \item \textbf{Setting}: This is when a player uses their hands to pass the ball high into the air for a teammate to spike.
        \end{itemize}
        \item \textbf{Analyze the woman's position}:
        \begin{itemize}
            \item The ball is above her head, and she appears to be in a position where she is about to use her hands to pass the ball.
            \item Her body posture suggests she is preparing to use her hands to control the ball, which is characteristic of bumping rather than setting.
        \end{itemize}
        \item \textbf{Consider the options}:
        \begin{itemize}
            \item Option (a) Bump: This aligns with her posture and the context of the game.
            \item Option (b) Set: This would require her to position herself differently and use her hands in a way that is not consistent with her current stance.
        \end{itemize}
    \end{enumerate}

    Given the analysis, the woman is more likely to bump the ball.

    \noindent\texttt{<think>} The woman is in a position to use her hands to control the ball, which is characteristic of bumping rather than setting.\texttt{</think>}
    \noindent\texttt{<answer>} \boxed{a} \texttt{</answer>}

    \end{mdframed}
\end{figure*}

\begin{figure*}[t]
    \centering
    \begin{mdframed}[
        frametitle={Perception Example 3},
        frametitlebackgroundcolor=lightyellow,
        backgroundcolor=gray!10,
        linecolor=gray,
        frametitlerule=true,
        frametitlefont=\color{black}\bfseries
    ]
        \centering
        \small
        \includegraphics[width=0.4\linewidth]{img/example/mmvp3.jpg} 
        \captionof{figure}{\centering Question: Can you see the caudal fin of the shark in the image?\\Options: (a) Yes (b) No\\Ground Truth: (a) Yes} 
        \flushleft

    \noindent\textbf{VLAA-Thinker-Qwen2.5VL-7B + DMLR:}
    To determine if the caudal fin of the shark is visible in the image, I need to analyze the orientation and positioning of the shark. The caudal fin is typically located at the rear end of the shark's body. In this image, the shark is shown from a side angle, and its tail is clearly visible extending towards the bottom right corner of the image. This indicates that the caudal fin can be seen.

    Therefore, based on the visibility of the tail in the image, the answer is (a) Yes. The answer is \boxed{a}.

    \vspace{1em}

    \noindent\textbf{VLAA-Thinker-Qwen2.5VL-7B:}

    \texttt{<think>}
    To determine if the caudal fin of the shark is visible in the image, I need to analyze the orientation and position of the shark. The caudal fin is typically located at the rear end of the shark's body. In this image, the shark is shown from a side angle, and the caudal fin is not directly visible due to the angle and the positioning of the shark's body. However, the tail area, which includes the caudal fin, can be inferred to be present based on the overall shape and structure of the shark.

    Given the options:
    (a) Yes
    (b) No

    The caudal fin is not directly visible in the image, but it can be inferred to be present based on the shark's anatomy and the angle of the image.

    Therefore, the correct answer is (b) No.
    \texttt{</think>}

    \noindent\texttt{<answer>} (b) No \texttt{</answer>}

    \end{mdframed}
\end{figure*}

\begin{figure*}[t]
    \centering
    \begin{mdframed}[
        frametitle={Perception Example 4},
        frametitlebackgroundcolor=lightyellow,
        backgroundcolor=gray!10,
        linecolor=gray,
        frametitlerule=true,
        frametitlefont=\color{black}\bfseries
    ]
        \centering
        \small
        \includegraphics[width=0.4\linewidth]{img/example/mmvp4.jpg} 
        \captionof{figure}{\centering Question: Is the flame of the match more round or thin?\\Options: (a) Round (b) Thin\\Ground Truth: (b) Thin} 
        \flushleft

    \noindent\textbf{VLAA-Thinker-Qwen2.5VL-7B + DMLR:}

    To determine whether the flame of the match is more round or thin, let's analyze the image. The flame appears elongated and narrow at the top, which suggests it is not round but rather thin in width. The base of the flame seems wider, but the overall shape of the flame is more elongated than circular.

    Therefore, based on the visual analysis, the flame of the match is more thin. The answer is \boxed{b}.

    \vspace{1em}

    \noindent\textbf{VLAA-Thinker-Qwen2.5VL-7B:}

    \texttt{<think>}
To determine whether the flame of the match is more round or thin, let's analyze the image. The flame appears to have a distinct shape where the base is wider and narrows towards the top, which is characteristic of a conical shape. This shape suggests that the flame is not uniformly round but rather has a tapering effect from bottom to top. Given this observation, the flame can be described as more round at its base and becoming thinner towards the top, but overall, it maintains a rounded appearance.

Therefore, based on the visual analysis, the flame of the match is more round.
    \texttt{</think>}

    \noindent\texttt{<answer>} \boxed{a} \texttt{</answer>}

    \end{mdframed}
\end{figure*}

\begin{figure*}[t]
    \centering
    \begin{mdframed}[
        frametitle={Reasoning Example 1},
        frametitlebackgroundcolor=lightyellow,
        backgroundcolor=gray!10,
        linecolor=gray,
        frametitlerule=true,
        frametitlefont=\color{black}\bfseries
    ]
        \centering
        \small
        \includegraphics[width=0.4\linewidth]{img/example/mathvision1.png} 
        \captionof{figure}{\centering Question: A beetle walks along the edges of a cube. Starting from point $P$ it first moves in the direction shown. At the end of each edge it changes the direction in which it turns, turning first right then left, then right etc. Along how many edges will it walk before it returns to point $P$?\\Ground Truth: 6} 
        \flushleft

    \noindent\textbf{Qwen2.5-VL-7B-Instruct + DMLR:}

    The beetle starts at point P and moves along the edges of a cube, changing direction according to the pattern: right, left, right. To determine how many edges it will walk before returning to point P, we need to trace its path.

    \begin{enumerate}
        \item The beetle starts at P.
        \item It moves right to an adjacent vertex.
        \item It turns left to another adjacent vertex.
        \item It turns right again to reach a third adjacent vertex.
        \item It continues this pattern until it returns to P.
    \end{enumerate}

    By examining the cube's structure, we can see that after 6 edges, the beetle will return to point P because each face of the cube has 4 edges, and the beetle covers all edges of one face before repeating the pattern on the next face.

    Thus, the beetle will walk along 6 edges before returning to point P.

    \vspace{1em}

    \noindent\textbf{Qwen2.5-VL-7B-Instruct:}

    To solve this problem, we need to track the beetle's path as it moves along the edges of the cube, changing its direction according to the specified pattern (right, left, right, etc.).

    \begin{enumerate}
        \item \textbf{Initial Position and Direction}: The beetle starts at point $P$ and moves in the direction indicated by the arrow, which is along one edge of the cube.
        \item \textbf{First Edge}: The beetle moves along the first edge and changes direction to the right.
        \item \textbf{Second Edge}: Moving right, the beetle now faces a new direction and continues along the next edge.
        \item \textbf{Third Edge}: After moving along the second edge, the beetle changes direction again to the left.
        \item \textbf{Fourth Edge}: Moving left, the beetle now faces another new direction and continues along the next edge.
        \item \textbf{Fifth Edge}: After moving along the third edge, the beetle changes direction to the right again.
        \item \textbf{Sixth Edge}: Moving right, the beetle now faces yet another new direction and continues along the next edge.
        \item \textbf{Seventh Edge}: After moving along the fourth edge, the beetle changes direction to the left once more.
        \item \textbf{Eighth Edge}: Moving left, the beetle now faces a new direction and continues along the next edge.
    \end{enumerate}

    At this point, the beetle has completed a full cycle around the cube, returning to point $P$.

    The beetle will have walked along 8 edges before it returns to point $P$. The answer is 8.

    \end{mdframed}
\end{figure*}

\begin{figure*}[p]
    \centering
    \begin{mdframed}[
        frametitle={Reasoning Example 2},
        frametitlebackgroundcolor=lightyellow,
        backgroundcolor=gray!10,
        linecolor=gray,
        frametitlerule=true,
        frametitlefont=\color{black}\bfseries
    ]
        \centering
        \small
        \includegraphics[width=0.4\linewidth]{img/example/mathvision2.png} 
        \captionof{figure}{\centering Question: In the adjoining plane figure, sides $AF$ and $CD$ are parallel, as are sides $AB$ and $EF$, and sides $BC$ and $ED$. Each side has length of 1. Also, $\measuredangle FAB = \measuredangle BCD = 60^\circ$. The area of the figure is\\Options: A. $\frac{\sqrt{3}}{2}$, B. $1$, C. $\frac{3}{2}$, D. $\sqrt{3}$, E. $2$\\Ground Truth: D. $\sqrt{3}$} 
        \flushleft

    \noindent\textbf{Qwen2.5-VL-7B-Instruct + DMLR:}

    The given figure is a hexagon with sides of equal length and specific angles. To find the area, we can break it down into simpler shapes like triangles and parallelograms.

    \begin{enumerate}
        \item \textbf{Identify the shape}: The hexagon can be divided into two congruent trapezoids by drawing a diagonal from one vertex to the opposite non-adjacent vertex.
        \item \textbf{Calculate the area of each trapezoid}:
        \begin{itemize}
            \item Each trapezoid has parallel sides of 1 unit (AF and CD), and the height can be calculated using the properties of the 60° angle.
            \item The height of the trapezoid can be found using the sine of 60° in a right triangle formed by dropping a perpendicular from one of the vertices to the base.
            \item The area of one trapezoid is $\frac{1}{2} \times (\text{sum of parallel sides}) \times \text{height}$.
        \end{itemize}
    \end{enumerate}

    The height $h$ of the trapezoid can be calculated as follows: $h = 1 \cdot \sin(60^\circ) = \frac{\sqrt{3}}{2}$.

    The area of one trapezoid is: $\text{Area of one trapezoid} = \frac{1}{2} \times (1 + 1) \times \frac{\sqrt{3}}{2} = \frac{1}{2} \times 2 \times \frac{\sqrt{3}}{2} = \frac{\sqrt{3}}{2}$.

    Since there are two such trapezoids, the total area of the hexagon is: $\text{Total Area} = 2 \times \frac{\sqrt{3}}{2} = \sqrt{3}$.

    \vspace{1em}

    \noindent\textbf{Qwen2.5-VL-7B-Instruct:}

    To find the area of the given figure, we need to recognize that the shape can be divided into two congruent trapezoids. Let's break down the problem step by step:

    \begin{enumerate}
        \item \textbf{Identify the Shape}: The given figure is a hexagon with specific angle and side lengths. Since $AF$ is parallel to $CD$, $AB$ is parallel to $EF$, and $BC$ is parallel to $ED$, the hexagon can be divided into two congruent trapezoids by drawing a diagonal from $A$ to $D$.
        \item \textbf{Calculate the Area of One Trapezoid}: Each trapezoid is formed by two right triangles and a rectangle in the middle. The key angles are $60^\circ$ at $FAB$ and $BCD$. This means that the trapezoid can be split into a rectangle and two 30-60-90 triangles.
        \item \textbf{Properties of 30-60-90 Triangles}: In a 30-60-90 triangle, the sides are in the ratio $1 : \sqrt{3} : 2$. Here, the shorter leg (opposite the $30^\circ$ angle) is 1, so the longer leg (opposite the $60^\circ$ angle) is $\sqrt{3}$, and the hypotenuse is 2.
        \item \textbf{Area Calculation}:
        \begin{itemize}
            \item The height of each trapezoid is the same as the shorter leg of the 30-60-90 triangle, which is 1.
            \item The bases of the trapezoid are the sum of the shorter legs of the two 30-60-90 triangles, which is $1 + 1 = 2$.
            \item The area of one trapezoid is given by the formula for the area of a trapezoid: $\text{Area} = \frac{1}{2} \times (\text{Base}_1 + \text{Base}_2) \times \text{Height}$.
            \item Substituting the values, we get: $\text{Area} = \frac{1}{2} \times (1 + 1) \times 1 = \frac{1}{2} \times 2 \times 1 = 1$.
        \end{itemize}
        \item \textbf{Total Area}: Since the hexagon is composed of two such congruent trapezoids, the total area is $1 + 1 = 2$.
    \end{enumerate}

    Therefore, the area of the figure is \boxed{2}.

    \end{mdframed}
\end{figure*} 

\begin{figure*}[p] 
    \centering
    \begin{mdframed}[
        frametitle={Reasoning Example 3 (Part 1)},
        frametitlebackgroundcolor=lightyellow,
        backgroundcolor=gray!10,
        linecolor=gray,
        frametitlerule=true,
        frametitlefont=\color{black}\bfseries
    ]
        \centering
        \small
        \includegraphics[width=0.4\linewidth]{img/example/mathvision3.png} 
        \captionof{figure}{\centering Question: The solid shown in the diagram has 12 regular pentagonal faces... (content omitted for brevity) ... Total of the numbers written on the solid?\\Ground Truth: 50} 
        \flushleft

    \noindent\textbf{VLAA-Thinker-Qwen2.5VL-7B + DMLR:}

    To solve this problem, we need to determine the number of triangular and square faces on the solid. Given that there are 12 pentagonal faces, each surrounded by 5 square faces, we can calculate the total number of square faces involved in these connections.

    Each pentagonal face is connected to 5 square faces, so for 12 pentagonal faces, the total count of square faces from pentagons alone would be $12 \times 5 = 60$. However, since each square face is shared between two pentagons, the actual number of unique square faces is half of this, which is $60 / 2 = 30$.

    Next, let's denote the number of triangular faces as $T$ and the number of square faces as $S$. We know: $S = 30$.
    The total number of faces is given by the sum of pentagonal, triangular, and square faces: $12 + T + 30 = F$, where $F$ is the total number of faces. Since each triangular face is surrounded by 3 square faces, and each square face is counted twice (once per triangle and once per pentagon), we can use the fact that each square face is part of 3 triangles or 2 pentagons to find $T$.

    Given the structure, we can derive: $3T = 2 \times 30 \implies 3T = 60 \implies T = 20$.

    Thus, the total number of faces $F$ is: $F = 12 + 20 + 30 = 62$.

    Now, calculating the total sum of numbers written on the solid:
    \begin{itemize}
        \item Each pentagonal face contributes $5$: $12 \times 5 = 60$
        \item Each triangular face contributes $1$: $20 \times 1 = 20$
        \item Each square face contributes $-1$: $30 \times -1 = -30$
    \end{itemize}

    Summing these contributions: $60 + 20 - 30 = 50$.

    Therefore, the total of the numbers written on the solid is $50$.

    \end{mdframed}
\end{figure*}

\begin{figure*}[t] 
    \centering
    \begin{mdframed}[
        frametitle={Reasoning Example 3 (Part 2)}, 
        frametitlebackgroundcolor=lightyellow,
        backgroundcolor=gray!10,
        linecolor=gray,
        frametitlerule=true,
        frametitlefont=\color{black}\bfseries
    ]
    \flushleft
    \small

    \noindent\textbf{VLAA-Thinker-Qwen2.5VL-7B:}

    \texttt{<think>}
    To solve this problem, we need to determine the number of each type of face (pentagon, triangle, and square) on the solid and then sum the values written on these faces according to the given rules.

    \begin{enumerate}
        \item \textbf{Identify the types of faces and their properties}:
        \begin{itemize}
            \item The solid has 12 pentagonal faces.
            \item Let $T$ be the number of triangular faces.
            \item Let $S$ be the number of square faces.
        \end{itemize}
        \item \textbf{Use the given information about the arrangement of faces}:
        \begin{itemize}
            \item Each pentagonal face is surrounded by 5 square faces.
            \item Each triangular face is surrounded by 3 square faces.
        \end{itemize}
        \item \textbf{Set up equations based on the surrounding faces}:
        \begin{itemize}
            \item Since each pentagonal face is surrounded by 5 square faces, and there are 12 pentagonal faces, the total contribution from pentagons to the count of square faces is $12 \times 5 = 60$.
            \item Since each triangular face is surrounded by 3 square faces, the total contribution from triangular faces to the count of square faces is $3T$.
        \end{itemize}
        \item \textbf{Since each square face is counted multiple times (once for each face it touches), we can set up the equation for the total number of square faces}:
        $60 + 3T = 4S$.
        This equation comes from the fact that each square face is part of one pentagonal face and $T$ triangular faces, contributing to the total count of square faces.
        \item \textbf{Determine the number of triangular faces $T$}: We know the total number of faces is $12 + T + S$. However, since each triangular face contributes 3 square faces and each pentagonal face contributes 5 square faces, we can use the relationship derived: $60 + 3T = 4S$.
        \item \textbf{Solve for $S$ in terms of $T$}: $4S = 60 + 3T \implies S = \frac{60 + 3T}{4}$.
        \item \textbf{Substitute back into the total face count equation}: $12 + T + S = 12 + T + \frac{60 + 3T}{4} = 12 + T + 15 + \frac{3T}{4} = 27 + \frac{7T}{4}$.
        \item \textbf{Solve for $T$ such that $S$ is an integer}:
        \begin{itemize}
            \item For $S$ to be an integer, $\frac{60 + 3T}{4}$ must be an integer. Testing values: $60 + 3T \equiv 0 \pmod{4} \implies 3T \equiv -60 \pmod{4} \implies 3T \equiv 0 \pmod{4} \implies T \equiv 0 \pmod{4}$.
            \item Trying $T = 4$: $S = \frac{60 + 3 \times 4}{4} = \frac{72}{4} = 18$.
        \end{itemize}
        \item \textbf{Verify the solution}:
        \begin{itemize}
            \item Total faces: $12 + 4 + 18 = 34$
            \item Square faces: $60 + 3 \times 4 = 72$ divided by 4: $18$
        \end{itemize}
        \item \textbf{Calculate the total value}:
        \begin{itemize}
            \item Pentagons: $12 \times 5 = 60$
            \item Triangles: $4 \times 1 = 4$
            \item Squares: $18 \times (-1) = -18$
            \item Total: $60 + 4 - 18 = 46$
        \end{itemize}
    \end{enumerate}

    Thus, the total of the numbers written on the solid is \boxed{46}.
    \texttt{</think>}

    \noindent\texttt{<answer>} \boxed{46} \texttt{</answer>}

    \end{mdframed}
\end{figure*}

\begin{figure*}[p]
    \centering
    \begin{mdframed}[
        frametitle={Reasoning Example 4 (Part 1)},
        frametitlebackgroundcolor=lightyellow,
        backgroundcolor=gray!10,
        linecolor=gray,
        frametitlerule=true,
        frametitlefont=\color{black}\bfseries
    ]
        \centering
        \small
        \includegraphics[width=0.4\linewidth]{img/example/mathvision4.png} 
        \captionof{figure}{\centering Question: We want to paint each square in the grid with the colours P, Q, R and S, so that neighbouring squares always have different colours. (Squares which share the same corner point also count as neighbouring.) Some of the squares are already painted. In which colour(s) could the grey square be painted?\\Options: A. only Q, B. only R, C. only S, D. either R or S, E. it is not possible.\\Ground Truth: D. either R or S} 
        \flushleft

    \noindent\textbf{VLAA-Thinker-Qwen2.5VL-7B + DMLR:}

    To determine the color of the grey square, we need to analyze the constraints given by the problem. The grid is filled with the colors P, Q, R, and S such that no two adjacent squares (including diagonally adjacent ones) share the same color.

    \begin{enumerate}
        \item \textbf{Identify Neighboring Squares}:
        \begin{itemize}
            \item The grey square has neighbors in the following positions:
            \item Top-left: Q
            \item Top-right: 
            \item Bottom-left: 
            \item Bottom-right: 
        \end{itemize}
        \item \textbf{Analyze the Top-Left Neighbor}:
        \begin{itemize}
            \item The top-left neighbor is already colored Q.
            \item Therefore, the grey square cannot be Q.
        \end{itemize}
        \item \textbf{Consider Possible Colors for the Grey Square}:
        \begin{itemize}
            \item Since the grey square cannot be Q, it must be either R or S.
            \item We need to check if both R and S are possible without violating the rule that no two adjacent squares can have the same color.
        \end{itemize}
        \item \textbf{Check for Validity of R}:
        \begin{itemize}
            \item If the grey square were R, then the top-right square would need to be a different color from R and Q. This leaves P as the only valid option for the top-right square since it must differ from both R and Q.
        \end{itemize}
        \item \textbf{Check for Validity of S}:
        \begin{itemize}
            \item If the grey square were S, then the bottom-right square would need to be a different color from S and Q. This leaves P as the only valid option for the bottom-right square since it must differ from both S and Q.
        \end{itemize}
        \item \textbf{Conclusion}:
        \begin{itemize}
            \item Both R and S are valid options for the grey square because they do not violate the adjacency rules when considering their neighboring squares.
        \end{itemize}
    \end{enumerate}

    Therefore, the grey square could be painted either R or S.

    \end{mdframed}
\end{figure*}

\begin{figure*}[t]
    \centering
    \begin{mdframed}[
        frametitle={Reasoning Example 4 (Part 2)},
        frametitlebackgroundcolor=lightyellow,
        backgroundcolor=gray!10,
        linecolor=gray,
        frametitlerule=true,
        frametitlefont=\color{black}\bfseries
    ]
        \flushleft
        \small

    \noindent\textbf{VLAA-Thinker-Qwen2.5VL-7B:}

    \texttt{<think>}
    To determine the color of the grey square, we need to consider the constraints given: no two neighboring squares can have the same color, including those that share a corner.

    \begin{enumerate}
        \item The square directly above the grey square is already colored 'Q'.
        \item The square to the left of the grey square is colored 'R'.
        \item The square below the grey square is colored 'Q'.
    \end{enumerate}

    Given these constraints:
    \begin{itemize}
        \item The grey square cannot be 'Q' because it shares a corner with the square above it which is 'Q'.
        \item The grey square cannot be 'R' because it shares a corner with the square to the left which is 'R'.
        \item The grey square cannot be 'P' because there is no information suggesting 'P' is used elsewhere in a way that would conflict with the rules for the grey square.
    \end{itemize}

    Therefore, the only remaining color for the grey square is 'S'.
    \texttt{</think>}

    \noindent\texttt{<answer>} \boxed{S} \texttt{</answer>}

    \end{mdframed}
    \label{fig8}
\end{figure*}

\section{Prompt}
\label{g}
In this section, we provide the complete set of prompts used in our experiments. These prompts cover all evaluation settings, including reasoning, perception, visual-grounding analysis, confidence-based studies, and ablation configurations. By releasing the full prompt formulations, we aim to ensure that every component of our experimental pipeline is transparent and fully reproducible, enabling researchers to faithfully replicate our results and conduct further analysis based on the same standardized inputs.

\label{prompt}
\begin{figure*}[t]
\centering
\begin{tcolorbox}[
    width=\textwidth,            
    colback=gray!10,
    colframe=gray,
    coltitle=white,
    fonttitle=\bfseries,
    title=Reasoning Chains Correct Judge,
    attach boxed title to top left,
    boxed title style={
        colback=gray!50!black,
        boxrule=0pt,
        arc=0pt,
        outer arc=0pt,
        top=3pt,
        bottom=3pt,
        left=4pt,
        right=4pt
    },
    enhanced,
    sharp corners,
    left=6pt,
    right=6pt,
    top=6pt,
    bottom=6pt,
    halign title=flush left,     
    before upper=\raggedright    
]
You are given a multimodal reasoning example.

-- The model receives an image and a question (possibly with answer options).\\
-- The model then produces a step-by-step reasoning chain, followed by a final answer.\\
-- You are also given the ground-truth correct answer.\\

Your task is to decide whether the model's final answer is correct or incorrect.

Please follow these rules:\\
1.~Ignore style and verbosity. Focus only on factual correctness.\\
2.~If the final answer matches the ground-truth answer, label it as \textbf{CORRECT}, even if some intermediate reasoning steps are slightly suboptimal.\\
3.~If the final answer does \textbf{NOT} match the ground-truth answer, label it as \textbf{INCORRECT}, even if parts of the reasoning look plausible.\\
4.~If the model outputs multiple conflicting answers, treat the final one as its final answer.\\

Return your decision in the following JSON format only:

\begin{verbatim}
{
  "label": "CORRECT" or "INCORRECT",
  "short_explanation": "<one-sentence>"
}
\end{verbatim}

\noindent QUESTION: \texttt{\{QUESTION\_TEXT\}}\\
OPTIONS (if any): \texttt{\{OPTION\_TEXT\}}\\
GROUND-TRUTH ANSWER: \texttt{\{GT\_ANSWER\}}\\
MODEL REASONING CHAIN: \texttt{\{MODEL\_CHAIN\}}\\
MODEL FINAL ANSWER: \texttt{\{MODEL\_FINAL\_ANSWER\}}

\end{tcolorbox}
\end{figure*}

\begin{figure*}[t]
\centering
\begin{tcolorbox}[
    width=\textwidth,
    colback=gray!10,
    colframe=gray,
    coltitle=white,
    fonttitle=\bfseries,
    title=Faithfulness Reasoning Judge,
    attach boxed title to top left,
    boxed title style={
        colback=gray!50!black,
        boxrule=0pt,
        arc=0pt,
        outer arc=0pt,
        top=3pt,
        bottom=3pt,
        left=4pt,
        right=4pt
    },
    enhanced,
    sharp corners,
    left=6pt,
    right=6pt,
    top=6pt,
    bottom=6pt,
    halign title=flush left,
    before upper=\raggedright
]
You are given a multimodal reasoning example.\par

The model receives:\par
- An image (you can refer to its description below), and\par
- A textual question (possibly with answer options).\par

The model then produces a step-by-step reasoning chain and a final answer.\par

Your task is to judge whether the reasoning chain is FAITHFUL to the information
in the question and image description, or UNFAITHFUL.\par

Definitions:\par
- A FAITHFUL chain explicitly uses relevant information from the question and image
  (or its description), and each step logically contributes to deriving the final answer.\par
- An UNFAITHFUL chain either:\par
  * ignores critical visual/textual details,\par
  * relies on generic templates or guesses,\par
  * contains logically inconsistent steps, or\par
  * gives a correct answer but with reasoning that does not match the actual evidence.\par

Instructions:\par
1. Check whether the reasoning steps are grounded in the given question and image description.\par
2. Check whether the logical steps are coherent and consistent with each other.\par
3. A chain can be UNFAITHFUL even if the final answer happens to be correct by coincidence.\par

Return your judgment in the following JSON format only:\par

\begin{verbatim}
{
  "label": "FAITHFUL" or "UNFAITHFUL",
  "short_explanation": "<one-sentence justification>"
}
\end{verbatim}

QUESTION:\par
\texttt{\{QUESTION\_TEXT\}}\par

IMAGE DESCRIPTION (if available):\par
\texttt{\{IMAGE\_DESCRIPTION\}}\par

MODEL REASONING CHAIN:\par
\texttt{\{MODEL\_CHAIN\}}\par

MODEL FINAL ANSWER:\par
\texttt{\{MODEL\_FINAL\_ANSWER\}}

\end{tcolorbox}
\end{figure*}

\begin{figure*}[t]
\centering
\begin{tcolorbox}[
    width=\textwidth,
    colback=gray!10,
    colframe=gray,
    coltitle=white,
    fonttitle=\bfseries,
    title=Visual Hallucination Judge,
    attach boxed title to top left,
    boxed title style={
        colback=gray!50!black,
        boxrule=0pt,
        arc=0pt,
        outer arc=0pt,
        top=3pt,
        bottom=3pt,
        left=4pt,
        right=4pt
    },
    enhanced,
    sharp corners,
    left=6pt,
    right=6pt,
    top=6pt,
    bottom=6pt,
    halign title=flush left,
    before upper=\raggedright
]
\small

You are given:\par
- An image,\par
- A question about the image, and\par
- Several reasoning statements produced by a model that refer to visual content.\par

Your task is to check whether each visual statement is supported by the image
(or its description), or whether it hallucinates objects, attributes, or relations
that do not actually appear in the image.\par

Definitions:\par
- A NON-HALLUCINATED (grounded) statement is consistent with the image:
  mentioned objects, attributes, and relations are all present or at least plausible.\par
- A HALLUCINATED statement mentions objects, attributes, numbers, or spatial relations
  that contradict the image or clearly do not appear in it.\par

Instructions:\par
1. Carefully compare each visual statement with the image (or its description).\par
2. For each statement, decide whether it is HALLUCINATED or NON-HALLUCINATED.\par
3. Ignore minor wording differences; focus on factual consistency.\par

Return your output in the following JSON format:\par

\begin{verbatim}
{
  "results": [
    {
      "statement_id": 1,
      "label": "HALLUCINATED" or "NON-HALLUCINATED",
      "short_explanation": "<short reason>"
    },
    {
      "statement_id": 2,
      "label": "...",
      "short_explanation": "..."
    }
    ...
  ]
}
\end{verbatim}

--------------------\par
QUESTION:\par
\texttt{\{QUESTION\_TEXT\}}\par

IMAGE DESCRIPTION (if available):\par
\texttt{\{IMAGE\_DESCRIPTION\}}\par

VISUAL STATEMENTS (from the model's reasoning):\par
1.~\texttt{\{STATEMENT\_1\}}\par
2.~\texttt{\{STATEMENT\_2\}}\par
3.~\texttt{\{STATEMENT\_3\}}\par
...\par

\end{tcolorbox}
\end{figure*}

\begin{figure*}[t]
\centering
\begin{tcolorbox}[
    width=\textwidth,
    colback=gray!10,
    colframe=gray,
    coltitle=white,
    fonttitle=\bfseries,
    title=Reasoning System Prompt (Ablation Study),
    attach boxed title to top left,
    boxed title style={
        colback=gray!50!black,
        boxrule=0pt,
        arc=0pt,
        outer arc=0pt,
        top=3pt,
        bottom=3pt,
        left=4pt,
        right=4pt
    },
    enhanced,
    sharp corners,
    left=6pt,
    right=6pt,
    top=6pt,
    bottom=6pt,
    halign title=flush left,
    before upper=\raggedright
]
\small

A conversation between a User and an Assistant.  
The User asks a question about an image, and the Assistant solves it.  
The Assistant first thinks through the reasoning process internally and then provides the final answer.

The reasoning process and the final answer must be enclosed within  
\texttt{<think>} \texttt{</think>} and \texttt{<answer>} \texttt{</answer>} tags, respectively.  
For example:\par
\texttt{<think>} reasoning process here \texttt{</think>}  
\texttt{<answer>} answer here \texttt{</answer>} \par

You are given:\par
- An image,\par
- A question about the image.\par

Your task:\par
1. Carefully analyze the image.\par
2. Identify objects, attributes, and spatial relations relevant to the question.\par
3. Perform step-by-step visual reasoning.\par
4. Give a concise and correct final answer.\par

Follow these rules:\par
- The reasoning inside \texttt{<think>} should be detailed and explicit.\par
- The final answer inside \texttt{<answer>} should be short and definitive.\par
- Do not mention the image analysis process outside of the \texttt{<think>} block.\par

--------------------\par
QUESTION:\par
\texttt{\{QUESTION\_TEXT\}}\par

IMAGE:\par
\texttt{\{IMAGE\_PLACEHOLDER\}}\par

\end{tcolorbox}
\end{figure*}

\begin{figure*}[t]
\centering
\begin{tcolorbox}[
    width=\textwidth,
    colback=gray!10,
    colframe=gray,
    coltitle=white,
    fonttitle=\bfseries,
    title=Fine-Grained Reasoning Prompt (visualization),
    attach boxed title to top left,
    boxed title style={
        colback=gray!50!black,
        boxrule=0pt,
        arc=0pt,
        outer arc=0pt,
        top=3pt,
        bottom=3pt,
        left=4pt,
        right=4pt
    },
    enhanced,
    sharp corners,
    left=6pt,
    right=6pt,
    top=6pt,
    bottom=6pt,
    halign title=flush left,
    before upper=\raggedright
]
\small

You are an Assistant that performs \textbf{fine-grained visual reasoning}.  
You are given an image and a question. Your task is to examine the image carefully,
identify all visually relevant details, and use them to answer the question.

Your reasoning process must:\par
- Attend to \textbf{small visual cues} such as colors, textures, shapes, numbers, and spatial relations.\par
- Explicitly verify each visual observation before relying on it.\par
- Avoid hallucinating any object, attribute, or relation not present in the image.\par
- Connect visual details step by step to form a coherent reasoning chain.\par

The reasoning process must be written inside \texttt{<think>} \texttt{</think>}.\par
The final short answer must be written inside \texttt{<answer>} \texttt{</answer>}.\par

Example format:\par
\texttt{<think>} detailed visual reasoning here \texttt{</think>}\par
\texttt{<answer>} final answer here \texttt{</answer>}\par

Instructions:\par
1. Begin by describing all key visual details relevant to the question.\par
2. Compare objects, count elements, check colors, sizes, and spatial layouts.\par
3. Use concrete grounding from the image before making any inference.\par
4. Produce the final answer only after completing the visual reasoning.\par

--------------------\par
QUESTION:\par
\texttt{\{QUESTION\_TEXT\}}\par

IMAGE:\par
\texttt{\{IMAGE\_PLACEHOLDER\}}\par

\end{tcolorbox}
\end{figure*}

\twocolumn

\clearpage

{
    \small
    \bibliographystyle{ieeenat_fullname}
    \bibliography{ref}
}


%% file: sec_arxiv/0_abstract.tex
\begin{abstract}
\vspace{10pt}
\centering
\textbf{\Large\textcolor{TitleText}{Abstract}}

AI agents are increasingly used to solve real-world tasks by reasoning over multi-turn user interactions and invoking external tools. However, applying reinforcement learning to such settings remains difficult: realistic objectives often lack verifiable rewards and instead emphasize open-ended behaviors; moreover, RL for multi-turn, multi-step agentic tool use is still underexplored; and building and maintaining executable tool environments is costly, limiting scale and coverage.
We propose \method, an RL framework that replaces verifiable outcome rewards with checklist rewards. \method decomposes each turn’s intended behavior into fine-grained binary criteria with explicit evidence grounding and structured metadata, turning open-ended judging into more stable classification-style decisions. To balance stability and informativeness, our method adopts a strategy of sparse reward assignment but dense evaluation criteria. Training is performed in a scalable LLM-simulated tool environment, avoiding heavy engineering for large tool sets.
Experiments show that \method consistently improves over supervised fine-tuning. Starting from a 8B Base model and training on an 8k-example RL dataset, \method improves over the SFT counterpart by \textbf{8} points on $\tau^2$-Bench, by \textbf{10} points on BFCL-V4, and by \textbf{12} points on ToolSandbox. The results match or even outperform similarly sized open-source baselines, including the judging model. \method thus provides a scalable recipe for optimizing multi-turn, multi-step tool-using agents without relying on verifiable rewards. Code provided by the open source community: \url{https://github.com/namezhenzhang/CM2-RLCR-Tool-Agent}
\vspace{10pt}
\end{abstract}

%% file: sec_arxiv/1_intro.tex
\section{Introduction}

\begin{figure*}[!t]
  \centering
  \includegraphics[width=\textwidth]{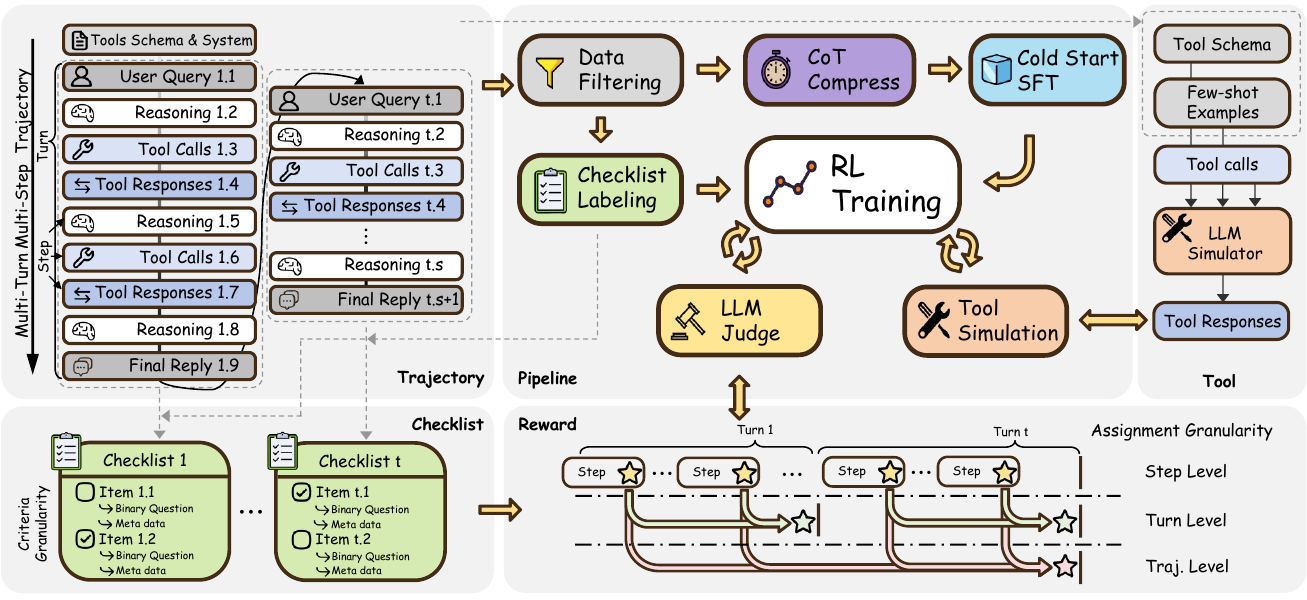}
  \caption{Overview of our \method. Starting from multi-turn, multi-step tool-use trajectories, we perform data filtering, CoT compression, and cold-start SFT, then annotate a per-turn checklist with evidence-grounded binary criteria and structured metadata. RL training is carried out in an LLM-simulated tool environment, where a LLM simulator produces tool responses and an LLM-as-a-Judge evaluates checklist items to compute rewards. The bottom panel contrasts dense criteria granularity with sparse reward assignment at different assignment granularities.}
  \label{fig:checklist_pipeline}
\end{figure*}

AI Agents are emerging as a promising paradigm for solving complex, real-world tasks~\citep{jimenez2023swe,phan2025humanity,wei2025browsecomp}. By reasoning and invoking external tools, such as search engines, databases, proprietary APIs, and compilers, an agent can interact with external environments to transcend the limitations of its parametric knowledge~\citep{jin2025search,jain2025r2e}. Unlike traditional question answering~\citep{kamalloo2023evaluating}, these agents require the ability to navigate \textbf{multi-turn} dialogues with users and execute \textbf{multi-step} reasoning with tool use~\citep{zhang2025survey}. However, training general-purpose agents to master such interactions through reinforcement learning (RL) remains a huge challenge.

\noindent Three primary limitations hinder current RL research in this domain. First, existing work largely relies on \textbf{verifiable rewards}~\citep{guo2025deepseek}. Typical setups supervise agents based on the rule-based correctness of final answers or the exact match of the tool execution trace against ground-truth~\citep{wei2025swe}. However, such signals are often not applicable in realistic, open-ended scenarios, where objectives may include asking clarifying questions, maintaining a helpful tone, or providing  suggestions~\citep{gunjal2025rubrics,liu2025openrubrics,viswanathan2025checklists}. Second, RL for multi-turn and multi-step interactions is underexplored. Most current works rely heavily on supervised fine-tuning (SFT) with synthetic data~\citep{qin2023toolllm}, simulated environment~\citep{xu2025epo,ALFWorld20} or RL limited to multi-step reasoning without multi-turn dynamics~\citep{yu2024steptool}. While these methods endow models with basic capabilities, they often struggle to generalize to unseen tools, extended horizons, and richer user interactions. Third, scaling tool-use RL is fundamentally constrained by tool environment construction.
Implementing tool APIs and maintaining reliable execution environments incurs substantial engineering overhead and makes it difficult to scale to large and diverse tools~\citep{liu2024toolace,ruan2023identifying}.

\noindent To address these challenges, we propose \method (\underline{C}hecklist Reward for \underline{M}ulti-turn \underline{M}ulti-step Agentic \underline{Tool} Use), an RL training framework for \textbf{multi-turn} and \textbf{multi-step} tool-use agent,  \textbf{without relying on rule-based verifiable rewards}.
The RL training is performed in a \textbf{scalable LLM-simulated tool environment} containing 5,000 tools. The workflow is illustrated in Figure~\ref{fig:checklist_pipeline}.



\noindent The core idea of \method is to replace the verifiable rewards with checklist rewards, decomposing the agent's intended behavior in each turn into fine-grained \textbf{binary} evaluation criteria, where each criterion is equipped with explicit evidence grounding, dependencies, and weights. This formulation turns open-ended judging into more stable classification-style decisions, while retaining interpretability and compositionality for complex objectives.

\noindent A central design question is how to trade off signal density and training stability under noisy tool simulations and LLM-based judging. We find that simply making rewards denser along the trajectory can amplify noise and destabilize optimization. \method therefore adopts a strategy: \textit{Sparse in assignment; Dense in criteria}. Rewards are assigned conservatively, while supervision remains informative. To study assignment granularity systematically, we instantiate three advantage estimation variants: trajectory-level, turn-level, and step-level. We also introduce a reward backfilling mechanism that attributes delayed checklist satisfaction to earlier critical steps when dependencies are met, improving credit assignment in long interactions.

\noindent To enable \textbf{scalable} training across diverse tools without heavy engineering, \method performs RL in an \textbf{LLM-simulated tool environment} containing 5,000+ tools. The simulator supports hybrid execution by replaying recorded tool I/O when available and falling back to LLM-based tool response simulation otherwise. This method enables large-scale, execution-free interaction while maintaining contextual consistency, thereby improving training robustness~\citep{liu2024toolace,ruan2023identifying}.

\noindent Empirically, \method yields significant improvements across multiple challenging benchmarks. 
Starting from a 8B base model and training on an 8k-example RL dataset, CM2 improves over the SFT counterpart by \textbf{$8$} points on $\tau^2$-Bench~\citep{yao2024tau,barres2025tau}, by 10 points on BFCL-V4~\citep{patil2025bfcl}, and by \textbf{$12$} points on ToolSandbox~\citep{lu2025toolsandbox}. The results match or even outperform similarly sized open-source baselines.
More importantly, \method enables robust reinforcement learning for agentic systems without requiring manual environment-specific reward engineering, demonstrating that Checklist rewards provide an effective and scalable supervision signal for training general-purpose agents capable of multi-turn, multi-step tool use, particularly invaluable in domains where explicit and verifiable rewards are unavailable, offering a practical pathway toward large-scale optimization of agentic tool use capabilities.

%% file: sec_arxiv/2_related_work.tex
\section{Related Work}

\subsection{Reward for RL}
Recent advances have shifted from SFT toward RL to enhance the generalization and robustness of agent behavior. A dominant paradigm is Reinforcement Learning with Verifiable Rewards (RLVR)~\citep{guo2025deepseek}, which leverages deterministic signals to guide optimization. However, applying RLVR to open-ended problems remains challenging due to the absence of ground-truth verifiers. Traditionally, Reinforcement Learning from Human Feedback~\citep{schulman2017proximal,rafailov2023direct} addresses this limitation by training reward models on human preference data to provide scalar signals~\citep{hong2025think,mahan2024generative}. Yet these holistic scalar rewards are often opaque and insufficient for guiding complex multi-step reasoning. To overcome this issue, recent work has turned to criterion-based rewards. Frameworks such as Reinforcement Learning with Rubric-based Rewards~\citep{gunjal2025rubrics,liu2025openrubrics,pathak2025rubric} and Reinforcement Learning from checklist Feedback~\citep{viswanathan2025checklists} decompose instruction execution into fine-grained checklist items or criteria, which are then evaluated by LLMs serving as judges. These studies demonstrate that dense, structured feedback substantially outperforms opaque scalar rewards from standard reward models.


\subsection{Multi-Turn Multi-Step Agent RL}
The evolution from single-step to multi-turn, multi-step agent interactions poses significant challenges for state tracking and credit assignment in RL training. Recent benchmarks~\citep{lu2025toolsandbox,barres2025tau,patil2025bfcl,zhou2025m} emphasize the importance of stateful dynamics, requiring agents to maintain contextual consistency and execute coherent tool-calling sequences over extended horizons. While these benchmarks effectively evaluate multi-turn dialogue or multi-step reasoning capabilities, existing work largely treats these two aspects in isolation, with few studies using RL to simultaneously optimize the compositional complexity arising from multi-turn dialogue dynamics and multi-step tool-use trajectories. 
Recently, 
MUA-RL~\citep{zhao2025mua} first integrated LLM-simulated users into RL loops but relies on binary outcome rewards and optimizes on in-domain evaluation data, failing to address sparse reward challenges in long interactions. In contrast, \method employs fine-grained checklist rewards to explicitly reinforce correct intermediate steps, effectively mitigating credit assignment problems and enabling more robust dialogue policies and tool-use patterns.

\subsection{LLM-Simulated Tool Environments}
The fundamental limitation in extending RL to tool-use domains lies in the engineering overhead of maintaining real-world APIs~\citep{lu2025toolsandbox,jain2025r2e}. To address this challenge, LLM-based environment simulation has become the dominant paradigm. 
SynthAgent~\citep{wang2025adapting} proposes a fully synthetic supervision framework for web agents with trajectory optimization to enhance performance; 
ToolEmu~\citep{ruan2023identifying} demonstrates the effectiveness of LLM-simulated sandboxes in identifying risky behaviors, enabling safety evaluation without actual tool infrastructure. Simia~\citep{li2025simulating} shows that powerful LLMs can faithfully simulate environment feedback based on tool definitions and interaction history, while Generalist Tool Model (GTM)~\citep{ren2025gtm} introduces a specialized 1.5B parameter model to simulate the execution of over 20,000 tools.
In contrast, \method scales to arbitrary tools, enabling large-scale training across diverse domains and synthetic edge cases that improve robustness.

%% file: sec_arxiv/3_method.tex
\section{RL via Checklist Rewards for Agentic Tool Use }
\label{sec:cm2}
In this section, we introduce our \method method. We first formulate the problem of multi-turn and multi-step agentic tool calling in Section~\ref{sec:problem_formulation} and then define two dimensions of granularity in reward modeling for agentic tasks in Section~\ref{sec:reward_granularity}. Subsequently, Section~\ref{sec:checklist_reward_shaping} describes the shaping and labeling process of the Checklist rewards. Finally, we detail how to do RL training with Checklist rewards in Section~\ref{sec:checklist_rl}.

\subsection{Problem Formulation}
\label{sec:problem_formulation}
As shown in the upper left part of Figure~\ref{fig:checklist_pipeline}, we consider a \textbf{multi-turn and multi-step} dialogue $\mathcal{D}$ between a user $u$ and an agent $\pi_\theta$ equipped with a set of tools $\mathcal{T} = \{ T_1, T_2, \ldots, T_K \}$.
A dialogue is composed of multiple \textbf{turns}: $\mathcal{D} = \{ \tau_1, \tau_2, \ldots, \tau_L \}$, where each turn $\tau_t$ consists of a sequence of \textbf{steps}: $\tau_t = \{ \sigma_{t,1}, \sigma_{t,2}, \ldots, \sigma_{t,M_t} \}$. Each step $\sigma_{t,s}$ is categorized into one of three types:
\textbf{(1) User Query}, marking the initiation of a turn;
\textbf{(2) Agent Action}, which comprises:
(i) an internal \textit{Reasoning process} $z_{t,s}$ that precedes an action, and
(ii) an explicit \textit{action} $a_{t,s}$, which may be tool calls or a final reply;
\textbf{(3) Tool Responses}, which are the output returned by the tool invoked in the preceding agent action.

\noindent We employ \textbf{Interleaved Thinking}~\citep{xie2025interleaved} to maintain context, and keep the thinking process from previous turns. The dialogue context $h_{t,s}$ is defined as the complete observable history up to step $\sigma_{t,s}$: $h_{t,s} = \{\tau_1, \ldots, \tau_{t-1}\} \cup \{ \sigma_{t,1}, \ldots, \sigma_{t,s} \}$.
At each agent action step, the model first generates reasoning $z_{t,s} \sim \pi_\theta(z \mid h_{t,s})$, followed by an action $a_{t,s} \sim \pi_\theta(a \mid h_{t,s}, z_{t,s})$.
If the action $a_{t,s}$ is a tool call, the tool environment executes the selected tool $T_i$ with arguments and returns an observation $r^{\text{tool}}_{t,s}=T_i(a_{t,s})$, which is then appended to the history to form $h_{t,s+1}$.
If the action is a final reply, the current turn terminates, and any subsequent user query initiates a new turn.

\begin{figure}[!t]
  \centering
  \begin{tcolorbox}[
    colback=promptbg,
    colframe=black!20,
    boxrule=0.5pt,
    arc=8pt,
  ]
\begin{Verbatim}
{
    "id": "D3",
    "evidence": [{
        "turn": 1, "step": 2,
        "from": "assistant.final_reply",
        "snippet": "which exceeds your $500 target.\n\n### Recommendations for Budget-Friendly Alternatives:"
    }],
    "focus_on": "assistant.final_reply",
    "question": "Does the assistant propose alternative budget-friendly van options or adjustments instead of generating a caption?",
    "pass_condition": "The final reply offers at least one cost-lowering alternative (e.g., cheaper van, longer term, smaller vehicle) and does not proceed to caption/hashtags.",
    "failure_examples":[
        "Assistant generates caption/hashtags despite payment > $500",
        "Assistant provides no alternative options"
    ],
    "strictness": true,
    "dependency": ["D1"],
    "weight": 0.2
}
\end{Verbatim}
  \end{tcolorbox}
  \caption{Example of One Checklist Item}
  \label{pr:checklist_example}
\end{figure}

\subsection{Two Types of Granularity in Reward Modeling}
\label{sec:reward_granularity}
Before detailing our Checklist reward shaping, we define two orthogonal dimensions of reward granularity: \textit{Assignment Granularity} and \textit{Criteria Granularity}. These dimensions address two fundamental questions: \textit{where} rewards are assigned along the trajectory, and \textit{what} criteria are used for evaluation.

\begin{table}[!]
\centering
\small
\caption{Components of a checklist item.}
\label{tab:checklist_metadata}
\begin{tabular}{lp{0.8\columnwidth}}
\toprule
\textbf{Component} & \textbf{Description} \\ \midrule
\textbf{Evidence} &
Pointers to the specific segment(s) in the original trajectory that this item is annotated from. \\
\textbf{Focus} &
The step type this item targets (e.g., tool calls, reasoning, final reply, or tool response), to help the judge localize the relevant context. \\
\textbf{Question} &
A binary checklist question to be answered for this item. \\
\textbf{Pass/Fail} &
Explicit criteria defining when the item passes or fails. \\
\textbf{Strictness} &
A boolean flag (required\_for\_next\_turn) indicating whether this item must pass for the conversation to proceed to the next turn since user query is fixed. \\
\textbf{Dependency} &
Dependencies indicating whether this item can only be satisfied after other item(s) are satisfied. \\
\textbf{Weight} ($w$) &
The item's relative weight within a turn, with $\sum_i w_i = 1$. \\
\bottomrule
\vspace{-20pt}
\end{tabular}
\end{table}

\noindent \textbf{Assignment Granularity} refers to the credit assignment of reward signals across the sequence of outputs. This dimension distinguishes between sparse and dense reward signals. At the coarse-grained level, the reward is assigned to the final state of a trajectory, treating the entire sequence as a single unit of evaluation. In contrast, the fine-grained level distributes reward signals across intermediate steps to evaluate the incremental progress of the generation. 

\noindent \textbf{Criteria Granularity} concerns the specificity of the evaluative metrics. Coarse-grained evaluation is holistic, where the reward reflects a single judgment, such as task completion or correctness. Fine-grained criteria decompose evaluation into multiple sub-dimensions (e.g., helpfulness, harmfulness, accuracy), each weighted according to a specific rubric.

\noindent While increasing granularity in both dimensions theoretically provides denser signals, our empirical observations in agentic scenarios suggest a decoupled strategy. Due to the inherent noise in the environment, \textbf{coarse-grained assignment} yields a more stable training curve. Concurrently, \textbf{fine-grained criteria} deliver the essential, task-specific guidance required to navigate complex tool-use logic. Consequently, we adopt a strategy characterized as \textit{Sparse in assignment; Dense in criteria}.

\subsection{Checklist Reward Shaping}
\label{sec:checklist_reward_shaping}
In this section, we introduce the \textbf{Checklist-based Reward Shaping} that can provide two types of fine-grained reward signals for multi-turn and multi-step RL training for agentic tool use.

\noindent\textbf{Composition of the Checklist.} As shown in the bottom left of Figure~\ref{fig:checklist_pipeline}, for each turn $\tau_t$, we label a \textit{Checklist} $\Gamma_{t}$ that contains several items $\{\gamma_1, \ldots, \gamma_{N_t}\}$. The annotator LLM is prompted to decompose the agent's intended behavior in each turn into multiple fine-grained subtasks. Each subtask, which is called a Checklist item, has one binary question and is enriched with detailed metadata that defines its semantics and constraints as shown in Table~\ref{tab:checklist_metadata}. The example of one Checklist item is illustrated in Figure~\ref{pr:checklist_example}.

\paragraph{Why Checklist Rewards?}
Checklist formulates each criterion as a \emph{binary} pass/fail decision with explicit evidence and conditions, turning LLM judging from open-ended scoring (regression) into a more \textbf{stable} and easy classification-style evaluation.
This substantially reduces judge randomness; otherwise, small stochastic score differences can be amplified by per-batch return or advantage normalization in RL, changing within-batch rankings and leading to unstable or even contradictory gradients~\citep{viswanathan2025checklists，jin2025your}.
Besides, this structured metadata ensures that the Checklist is \textbf{interpretable}, allowing automated and consistent evaluation across turns with less noise.

\paragraph{Post-hoc Checklist Annotation.} In practice, we label the Checklist by \emph{post-hoc} structuring an existing multi-turn and multi-step tool use trajectory rather than from scratch.
For each turn, we prompt an LLM to (i) infer the turn-level intent and required outcomes from the user query and the assistant/tool traces, and (ii) decompose them into a concise set of \textbf{binary, observable} Checklist items grounded in the trajectory. Each trajectory only costs approximately \$0.1 on average, making it practical to scale checklist labeling to large datasets without significant overhead compared with training costs and manual annotation.
The prompt and annotation details are provided in Appendix~\ref{sec:prompt_checklist_annotation}.

\paragraph{Rollout and Reward Computation.}
During rollout, at each step within turn $\tau_t$, we query a judge LLM with the trajectory prefix (history so far) together with the checklist items for that turn. The judge returns a Boolean label for each item, indicating whether it is currently satisfied by the partial trajectory.
After the agent produces the final user-visible response for the turn, we enforce the strictness constraints: if all strictness items are satisfied, we issue the next user query from the reference trajectory; otherwise, we terminate the rollout early.

\subsection{Checklist-based RL Optimization}
\label{sec:checklist_rl}
RL algorithm based on Group Relative Policy Optimization (GRPO) are typically formulated around outcome rewards.
However, our Checklist-based framework enables the extraction of dense reward signals down to the individual step level. To systematically investigate the impact of \textit{Assignment Granularity}, we instantiate three distinct advantage estimation variants: (i) Trajectory-level, (ii) Turn-level, and (iii) Step-level. These variants differ primarily in how to assign the reward and calculate the advantage accordingly.
\vspace{-2pt}
\subsubsection{Checklist-based Reward}
Let $x_s$ denote the state before step $s$ and $x_{s+1}$ the state after step $s$.
For dialogue $i$, turn $t$, and checklist item $c$, let $\mathrm{Sat}^{(i)}_{t,c}(x_s)\in\{0,1\}$ denote whether $\gamma^{(i)}_{t,c}$ is satisfied in state $x_s$.
Let 
\begin{equation}
\mathrm{Dep}_{t,c} = \{\, c' \mid \gamma_{t,c'} \text{ is a dependency (prerequisite) of } \gamma_{t,c} \,\},
\end{equation}
be the set of dependency items of $\gamma_{t,c}^{(i)}$.
Once $\gamma^{(i)}_{t,c}$ switches from unsatisfied to satisfied at step $s$, and all its dependencies are already satisfied in the pre-step state $x_s$, we assign a binary reward to that step:
\begin{equation}
r^{(i)}_{t,s,c}=\mathbf{1}\Big[
\underbrace{\prod_{c' \in \mathrm{Dep}_{t,c}} \mathrm{Sat}^{(i)}_{t,c'}(x_s) = 1}_{\text{all deps.\ satisfied in } x_s} \land\ 
\underbrace{\mathrm{Sat}^{(i)}_{t,c}(x_s)=0}_{\text{unsatisfied in } x_s} \land\
\underbrace{\mathrm{Sat}^{(i)}_{t,c}(x_{s+1})=1}_{\text{satisfied in } x_{s+1}}
\Big].
\end{equation}

\noindent Since satisfying an item may require multiple steps, we further \emph{backfill} the reward
to every earlier step where all the dependencies were already satisfied. The backfilled reward is defined as
\begin{equation}
\tilde r^{(i)}_{t,s,c}=\mathbf{1}\Big[
\underbrace{\prod_{c' \in \mathrm{Dep}_{t,c}} \mathrm{Sat}^{(i)}_{t,c'}(x_s) = 1}_{\text{all deps. satisfied before \(s\)}} \land\ 
\underbrace{\mathrm{Sat}^{(i)}_{t,c}(x_s)=0}_{\text{unsatisfied before \(s\)}} \land\ 
\underbrace{\exists\, u \ge s \ \text{s.t.}\ \mathrm{Sat}^{(i)}_{t,c}(x_{u+1})=1}_{\text{satisfied after \(s\)}}
\Big].
\end{equation}
Note that we only use \emph{backfilled} reward in step-level advantage.

\subsubsection{Trajectory-level Advantage}
Given a dialogue (rollout) \(D^{(i)} = \{ \tau_1, \ldots, \tau_{L^{(i)}} \}\), we first aggregate all Checklist-based rewards across turns, steps, and items as
\begin{equation}
R^{(i)} \;=\; \frac{1}{L^{(i)}} \sum_{t=1}^{L^{(i)}} \sum_{s} \sum_{c} w_{t,c} \cdot r^{(i)}_{t,s,c},
\end{equation}
where $s$ ranges over steps in turn $t$ and $c$ ranges over checklist items for turn $t$ and
\(R^{(i)} \in [0,1]\) since $\sum_{s} r^{(i)}_{t,s,c} \le 1 $ (it only flips once) and $\sum_{s} \sum_{c} w_{t,c} \cdot r^{(i)}_{t,s,c} \le \sum_{c} w_{t,c} =1$.
For the group of \(G\) rollouts of the same prompt, we define the trajectory-level advantage as
\begin{equation}
A^{(i)}_{\mathrm{traj}} \;=\; \frac{R^{(i)} \;-\; \text{mean}(\{R^{(i)}\}_{i=1}^{G})}{F_{\text{norm}}(\{R^{(i)}\}_{i=1}^{G})}.
\end{equation}

\subsubsection{Turn-level Advantage}
To get the turn-level advantage, we aggregate Checklist-based rewards \emph{within} each turn. For dialogue $i$ and turn $t$, we define the turn reward as
\begin{equation}
R^{(i)}_{t} \;=\; \sum_{s} \sum_{c} w_{t,c} \cdot r^{(i)}_{t,s,c} ,
\end{equation}
where $s$ ranges over steps in turn $t$ and $c$ ranges over checklist items for turn $t$ and \(R^{(i)}_{t} \in [0,1]\).
Given a group of \(G\) rollouts of the same question, we compute a turn-level GRPO advantage as
\begin{equation}
A^{(i)}_{\mathrm{turn},t} \;=\; \frac{R^{(i)}_{t} \;-\; \mathrm{mean}\!\left(\{R^{(i)}_{t}\}_{i=1}^{G}\right)}{F_{\text{norm}}\!\left(\{R^{(i)}_{t}\}_{i=1}^{G}\right)}.
\end{equation}

\subsubsection{Step-level Advantage}
For the step-level reward baseline, we first calculate a baseline in one group satisfy a certain Checklist item:
\begin{equation}
b_{t,c} \;=\; \frac{1}{G} \sum_{i=1}^{G} \mathbb{I} \big[ \exists s' \ \text{s.t.} \ r^{(i)}_{t,s',c} = 1 \big].
\end{equation}

\noindent At step $(t,s)$ in rollout $i$, multiple checklist items may be applicable simultaneously. We first compute an item-wise step advantage:
\begin{equation}
A^{(i)}_{t,s,c} \;=\; \frac{\tilde r^{(i)}_{t,s,c} \;-\; b_{t,c}}{F_{\text{norm}}(\{\mathbb{I} \big[ \exists s' \ \text{s.t.} \ r^{(i)}_{t,s',c} = 1 \big]\}_{i=1}^{G})},
\end{equation}
and then aggregate them using the checklist weights:
\begin{equation}
\begin{aligned}
A^{(i)}_{\mathrm{step},t,s}
&=
\frac{\sum_{c \in \mathcal{E}^{(i)}_{t,s}} w_{t,c}\,A^{(i)}_{t,s,c}}
{\sum_{c \in \mathcal{E}^{(i)}_{t,s}}w_{t,c}}.
\end{aligned}
\end{equation}
Here
$
\mathcal{E}^{(i)}_{t,s} \;=\; \bigl\{ c \,\big|\, \prod_{c' \in \mathrm{Dep}_{t,c}} \mathrm{Sat}^{(i)}_{t,c'}(x_s) = 1 \land \mathrm{Sat}^{(i)}_{t,c}(x_s) = 0 \bigr\}
$
denotes the set of checklist items that are eligible to be satisfied at step $s$ (i.e., all dependencies are already satisfied and item $c$ is not yet satisfied).

%% file: sec_arxiv/4_pipeline.tex
\section{Training Pipeline}
\label{sec:training_pipeline}
In this section, we outline the training pipeline of \method, which encompasses data filtering, Chain-of-Thought (CoT) compression, cold-start SFT, checklist labeling, tool simulation, and RL training guided by an LLM-as-a-Judge. The overall workflow is illustrated in the top right of Figure~\ref{fig:checklist_pipeline}.

\subsection{Data Filtering}
We start from the tool-calling subset of the \textsc{nvidia/Nemotron-Post-Training-Dataset-v1} dataset~\citep{NemotronPostTrainingDatasetV1, bercovich2025llamanemotronefficientreasoningmodels}, which contains 310k synthetic tool-use dialogues spanning single-turn, multi-turn, and multi-step settings across diverse domains (e.g., shopping, financial analysis, and web search). Since all samples are distilled from an LLM, the data contains substantial noise. We therefore apply a two-stage filtering pipeline to ensure quality: (1) \textbf{Rule-based filtering} removes examples with structural and formatting violations (criteria in Appendix~\ref{app:rule_base_filtering_criteria});
(2) \textbf{LLM-based filtering} uses GPT-5\citep{openai2025gpt5} to further discard samples with deeper semantic or reasoning errors. The prompt and details are provided in Appendix~\ref{sec:prompt_llm_based_filtering}.

We also conducted additional experiments on the APIGen-MT-5k dataset\citep{prabhakar2025apigen}, but we did not clean it.

\paragraph{Data statistics.}
Rule-based filtering reduces the dataset from 310k to 280k examples, and LLM-based filtering further narrows it to 30k high-quality samples. From this set, we randomly sample \textbf{8k examples} for cold-start SFT, and the remaining 22k form the candidate pool for RL, from which we additionally exclude simpler cases (e.g., single-turn or single-tool interactions) and retain another \textbf{8k complex multi-turn, multi-step dialogues} for RL training, with 500 held out for validation.

\subsection{CoT Compression and Cold Start}
Before finalizing the training sets, we compress the original chain-of-thought (CoT) to improve inference efficiency and reduce context length. Specifically, we use GPT-5 to rewrite the thinking content into a shorter form while preserving the key planning and decisions (prompt in Appendix~\ref{sec:prompt_shorten_thinking}). After compression, the resulting datasets are denoted as $\mathcal{D}_{\text{CS}}$ (cold-start SFT) and $\mathcal{D}_{\text{RL}}$ (RL training), respectively.

\noindent Finally, we fine-tune a 8B base model on the $\mathcal{D}{\text{CS}}$. Hyperparameters and other training details are provided in Appendix~\ref{app:cold_start_hyperparameters}.

\vspace{-5pt}
\subsection{Tool Simulation and LLM-as-a-Judge}
Because trajectories are synthetic, there is no executable environment available during RL. To avoid building and maintaining 5,000+ unique tools, we implement a hybrid tool simulator.
Upon a tool invocation, the simulator first performs an \textbf{exact match} against the original tool name and arguments; if matched, it returns the recorded tool response. Otherwise, we fall back to \textbf{LLM-based simulation}: we prompt an LLM with in-dialogue tool I/O exemplars as few-shot learning to generate a response that remains consistent with the trajectory context, enabling scalable, execution-free interaction. For \textbf{LLM-as-a-Judge} for checklist rewards, we prompt an LLM at each step to answer each question in checklist. Then we aggregate reward as in Section~\ref{sec:cm2}.
The judging prompt is provided in Appendix~\ref{sec:prompt_llm_judge}. We use a 30B model with 3B Activate parameters for both tool simulation and LLM-as-a-Judge, chosen to balance quality and throughput. Later experiments show that even a lightweight judge with merely 3B active parameters enables the model to attain highly competitive or even surpassing results.

\subsection{Checklist Labeling and RL Training}
Following Section~\ref{sec:checklist_reward_shaping}, we use GPT-5 to annotate a per-turn Checklist for each dialogue (prompts in Appendix~\ref{sec:prompt_checklist_annotation}). We then optimize from the cold-start SFT checkpoint using GRPO based on VeRL, and apply the multi-level advantage comparison described in Section~\ref{sec:checklist_rl}. The RL model is trained on 64 GPUs for 680 hours. Additional implementation details and hyperparameters are deferred to Appendix~\ref{app:rl_training_details}.

%% file: sec_arxiv/5_no_qwen_results.tex
\section{Results}


\begin{figure*}[!t]
  \centering
  \begin{subfigure}[t]{0.49\textwidth}
    \centering
    \includegraphics[width=\linewidth]{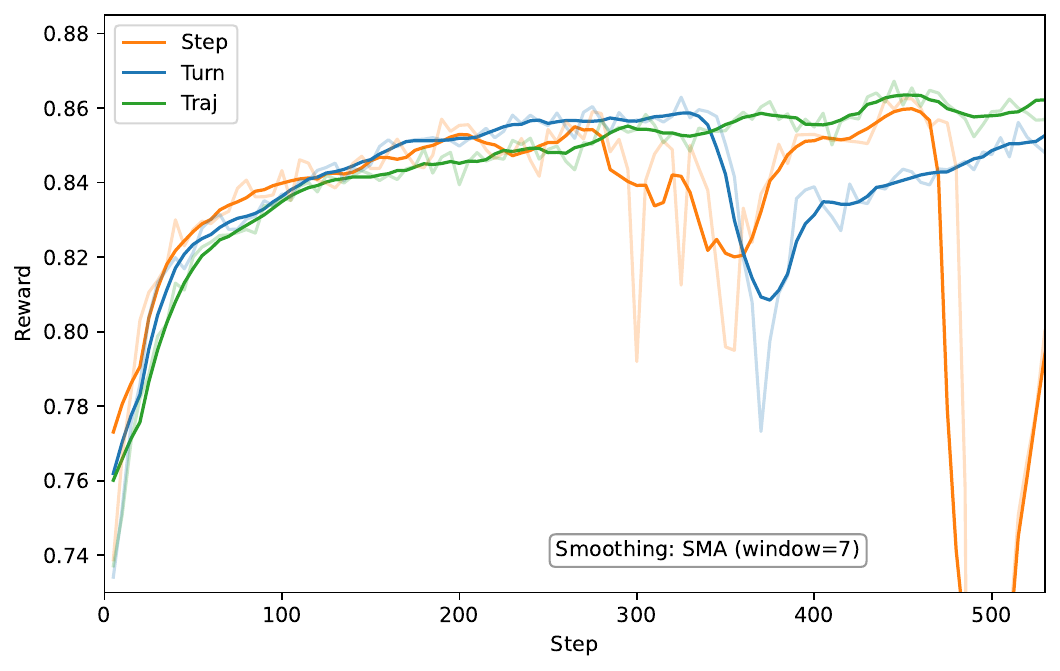}
    \caption{Reward curves on the validation set for different advantage assignment granularities.}
    \label{fig:allocation_comparison}
  \end{subfigure}
  \hfill
  \begin{subfigure}[t]{0.49\textwidth}
    \centering
    \includegraphics[width=\linewidth]{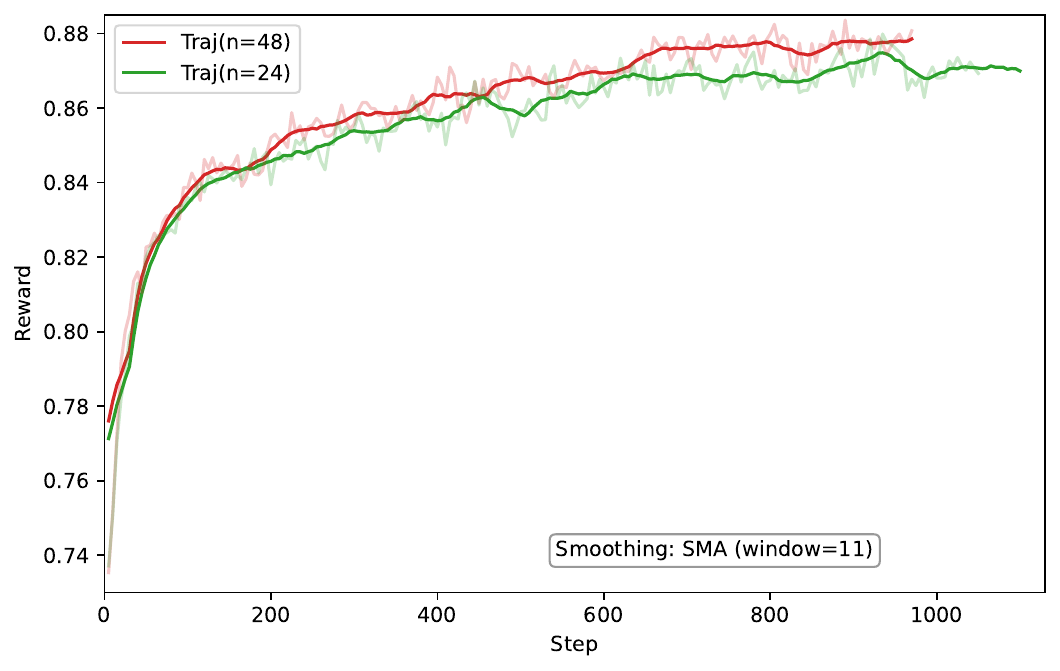}
    \caption{Reward curves on the validation set between different group sizes ($n$).}
    \label{fig:group_size_comparison}
  \end{subfigure}

  \caption{Comparison results under different settings.}
  \label{fig:two_in_one}
\end{figure*}

\subsection{Effect of Allocation Granularity}
Figure~\ref{fig:allocation_comparison} compares the reward curve on validation set under different assignment granularities. Finer-grained allocation yields faster early improvements: step-level advantages outperform turn-level, which in turn outperform trajectory-level in the initial phase.
As training continues, however, finer granularities exhibit earlier and more severe training collapse, while trajectory-level advantages remain more stable and continue to improve. 

\noindent We attribute this trade-off to noise amplification in agentic RL. Checklist rewards reduce judge variance by turning open-ended scoring into binary, evidence-grounded decisions, but they do not eliminate stochasticity. With finer-grained assignment, this residual noise enters optimization more frequently and can be amplified by group-relative normalization, yielding higher-variance or sometimes misleading gradients. This motivates our principle of \emph{Sparse in assignment; Dense in criteria}: we keep evaluation criteria fine-grained for informative supervision, while assigning rewards at coarser granularity to average out residual noise and improve stability.

\subsection{Effect of Group Size}
Figure~\ref{fig:group_size_comparison} shows the impact of group size $G$ (e.g., $G{=}24$ vs.\ $G{=}48$) with trajectory-level Checklist rewards. A larger group size consistently achieves higher rewards. Intuitively, for multi-turn, multi-step trajectories, increasing $G$ provides more samples for later turns, leading to a lower-variance advantage estimate more reliable gradient updates.








\subsection{Results on Benchmarks}

\newpage
\setlength{\tabcolsep}{4pt}
\begin{table}[!h]
\centering
\resizebox{0.5\textwidth}{!}{
\begin{tabular}{lcccc}
\toprule
\textbf{Model / Method} & \textbf{Airline} & \textbf{Retail} & \textbf{Telecom} & \textbf{Avg.} \\
\midrule
\rowcolor{gray!12}
\multicolumn{5}{c}{\textit{Open-source Baselines}} \\
30B-A3B-Instruct & 32.50 & 50.88 & 12.72 & 32.03 \\
8B-Thinking & 30.00 & 43.64 & 22.37 & 32.00 \\
\rowcolor{gray!12}
\multicolumn{5}{c}{\textit{Ours} (from \textbf{8B-Base})} \\
Cold-start SFT on $\mathcal{D}_{\text{CS}}$ & 25.50 & 18.42 & 11.84 & 18.59 \\
\quad $\hookrightarrow$ SFT on $\mathcal{D}_{\text{RL}}$ & 23.50 & 19.52 & 12.06 & 18.36 \\
\quad $\hookrightarrow$ RL on $\mathcal{D}_{\text{RL}}$ (\method) & 27.00 & 36.40 & 16.89 & \textbf{26.76} \\
\rowcolor{gray!12}
\multicolumn{5}{c}{\textit{Ours} (from 8B-Thinking)} \\
\quad SFT on $\mathcal{D}_{\text{In-domain}}$ & 30.00& 44.74 & 23.68 & 32.81 \\
\quad RL on $\mathcal{D}_{\text{In-domain}}$ (\method-$\tau^2$) & 33.00 & 54.17 & 37.00 & \textbf{41.39} \\
\bottomrule
\end{tabular}}
\caption{Results on the $\tau^2$-Bench benchmark. We run evaluation four times and report the average accuracy
We follow the default evaluation setting of $\tau^2$-Bench, except that we retain the thinking content for our models to align with our training setup.
We further conduct SFT and RL with a sythetic 5k in domain data of $\tau^2$-Bench. It shares the same tool functions on airline and retail subset while telecom is not covered.
}
\label{tab:tau2_bench}
\end{table}

\begin{table}[!h]
\centering
\resizebox{0.8\columnwidth}{!}{%
\begin{tabular}{l cccccccc}
\toprule
& \multicolumn{5}{c}{\textbf{Multi-Turn}} & \multicolumn{3}{c}{\textbf{Web Search}} \\
\cmidrule(lr){2-6} \cmidrule(lr){7-9}
\textbf{Model / Method}  & \textbf{Base} & \textbf{Miss Func} & \textbf{Miss Param} & \textbf{Long Ctx} & \textbf{Overall} & \textbf{Base} & \textbf{No Snippet} & \textbf{Overall} \\
\midrule
\rowcolor{gray!12}
\multicolumn{9}{c}{\textit{Open-source Baselines}} \\
30B-A3B-Instruct-2507  & 45.0 & 28.0 & 21.0 & 42.5 & \textit{34.25}  &	24.00 &	17.00 & 20.50\\
8B-Thinking & 42.5 & 38.5 & 31.5 & 35.5 & \textit{37.00} & 	19.00 & 11.00 & 15.00 \\
\rowcolor{gray!12}
\multicolumn{9}{c}{\textit{Ours} (from \textbf{8B-Base})} \\
Cold-start SFT on $\mathcal{D}_{\text{CS}}$  & 24.5 & 19.0 & 14.5 & 19.5 & \textit{19.37} & 18.00 & 10.00 & 14.00 \\
\quad $\hookrightarrow$ SFT on $\mathcal{D}_{\text{RL}}$  & 30.0 & 27.5 & 24.5 & 25.0 & \textit{26.75} & 18.00 & 9.00 & 13.50 \\
\quad $\hookrightarrow$ RL on $\mathcal{D}_{\text{RL}}$ (\method) & 44.5 & 32.0 & 35.0 & 34.5 & \textit{\textbf{36.50}} & 41.00 & 14.00 & \textbf{27.50} \\
\bottomrule
\end{tabular}%
}
\caption{Results on the BFCL-V4 benchmark (Multi-Turn and Web Search subset).
}
\label{tab:bfcl_results}
\end{table}

\begin{table}[!h]
\centering
\resizebox{\textwidth}{!}{%
\begin{tabular}{l c ccccccc cccccccc}
\toprule
\textbf{Model / Method} &
\multicolumn{7}{c}{\textbf{Scenario Categories}} &
\multicolumn{8}{c}{\textbf{Tool Augmentations}} & \textbf{Overall Score} $\uparrow$ \\
 \cmidrule(lr){2-8} \cmidrule(lr){9-16} \cmidrule(lr){17-17}
&
\textbf{STC} & \textbf{MTC} & \textbf{SUT} & \textbf{MUT} & \textbf{SD} & \textbf{C} & \textbf{II} &
\textbf{0-DT} & \textbf{3-DT} & \textbf{10-DT} & \textbf{AT} & \textbf{TNS} & \textbf{TDS} & \textbf{ADS} & \textbf{ATS} &  \\
\midrule
\rowcolor{gray!12}
\multicolumn{17}{c}{\textit{Open-source Baselines}} \\
30B-A3B-Instruct-2507 & 84.18 & 69.14 & 74.52 & 65.33 & 75.11 & 66.95 & 40.97 & 64.29 & 68.23 & 60.98 & 66.62 & 68.56 & 63.32 & 66.17 & 63.74 & 65.24 \\
8B-Thinking & 77.12 & 58.91 & 64.07 & 57.82 & 60.65 & 56.71 & 76.77 & 70.96 & 67.69 & 64.55 & 60.79 & 69.00 & 56.98 & 65.08 & 68.71 & 65.47 \\
\rowcolor{gray!12}
\multicolumn{17}{c}{\textit{Ours} (from \textbf{8B-Base})} \\
Cold-start SFT on $\mathcal{D}_{\text{CS}}$  & 71.65 & 47.89 & 54.91 & 45.72 & 63.08 & 45.93 & 69.97 & 55.44 & 56.68 & 56.03 & 53.86 & 60.34 & 55.81 & 52.82 & 58.53 & 56.19 \\
\quad $\hookrightarrow$ SFT on $\mathcal{D}_{\text{RL}}$  & 74.89 & 46.66 & 55.71 & 42.22 & 59.25 & 44.35 & 67.41 & 55.69 & 54.09 & 55.23 & 50.42 & 62.42 & 55.27 & 53.42 & 56.04 & 55.32 \\
\quad $\hookrightarrow$ RL on $\mathcal{D}_{\text{RL}}$ (\method) & 78.46 & 66.12 & 69.23 & 63.40 & 67.36 & 63.41 & 70.31 & 69.82 & 63.97 & 65.89 & 65.25 & 74.06 & 67.03 & 67.78 & 71.81 & \textbf{68.20} \\
\bottomrule
\end{tabular}%
}
\caption{Performance of ToolSandbox on various scenarios and tool augmentations.
Our models are trained from \textbf{8B-Base}; we do not report results for the base checkpoint since it is not instruction-tuned under this evaluation protocol. Here, $\mathcal{D}_{\text{CS}}$ denotes the 8k cold-start SFT set, and $\mathcal{D}_{\text{RL}}$ denotes the 8k complex multi-turn, multi-step RL training set.}
\label{tab:scenario_tool_breakdown}
\end{table}

We evaluate our proposed \method using our final configuration (trajectory-level advantage estimation with group size $G=48$) on three challenging multi-turn, multi-step tool-use benchmarks: $\tau^2$-Bench, BFCL-V4, and ToolSandbox. We compare against the SFT counterparts and open-source models of similar size.

\paragraph{$\tau^2$-Bench Benchmark.}
The results on $\tau^2$-Bench are summarized in Table~\ref{tab:tau2_bench}.
For each question, we run evaluation four times and report average accuracy. As shown in Table~\ref{tab:tau2_bench}, starting from an \textbf{8B base model}, our RL model outperforms SFT by over 8 points, demonstrating the effectiveness of \method. 
However, our RL training uses a maximum context length of 10k and up to 30 turns, whereas $\tau^2$-Bench can require $>$30k context and up to 200 turns. Under this mismatch, \method lags behind some open-source models such as an 30B instruct model with 3B activation parameters, and a 8B reasoning model.

To mitigate this, we further perform RL on an in-domain dataset with 5k data, which substantially improves average performance and surpasses the open-source baselines. Overall, these results indicate that \method is particularly effective when paired with in-domain data.

\paragraph{BFCL Benchmark.} 
Table~\ref{tab:bfcl_results} summarizes the results on BFCL-V4 (Multi-Turn and Web Search).
Overall, our RL model trained on $\mathcal{D}_{\text{RL}}$ (\method) substantially improves over SFT variants: on Multi-Turn, it achieves 36.50 overall accuracy, outperforming cold-start SFT and further SFT on $\mathcal{D}_{\text{RL}}$ by 10 points. 
On Web Search, RL also yields the best overall performance, improving over cold-start SFT and SFT on $\mathcal{D}_{\text{RL}}$ by 13.5 and 14 points, respectively.
Compared with open-source baselines, our RL model performs better than 30B-A3B-Instruct model (judging model) on Multi-Turn and is comparable to 8B-Thinking model, while significantly surpassing both baselines on Web Search.

\paragraph{ToolSandbox Benchmark.}
Table~\ref{tab:scenario_tool_breakdown} reports performance on ToolSandbox Benchmark.
RL on $\mathcal{D}_{\text{RL}}$ (\method) yields a large improvement over both SFT variants, increasing the overall score by more than 12 points.
It also improves consistently across nearly all scenario categories, with particularly notable gains on multi-turn and multi-tool settings.
Our RL model (\method) also outperforms the open-source models, including the judging model.

\paragraph{Summary.}
Our method consistently yields substantial gains over SFT, with improvements that are stable across benchmarks. Notably, the resulting policy matches and often surpasses the LLM-as-a-judge model on most evaluation measures, while remaining competitive with or exceeding similarly sized open-source baselines. We further find that a lightweight judge is sufficient to drive strong RL improvements, and the learned behavior generalizes well to previously unseen benchmarks.



%% file: sec_arxiv/7_discussion_conclusion.tex
\section{Discussion: Scaling Up}
There are several natural axes to scale up \method. First, we can increase the \emph{number of checklists per turn} by generating multiple, independently instantiated checklists for the same turn (e.g., with different paraphrases or decompositions). Aggregating their outcomes (e.g., averaging or majority voting) can further reduce residual stochasticity and improve robustness to occasional missing or ambiguous criteria, at the cost of additional judging compute.
Second, we can reduce judge noise more directly via \emph{majority vote} (or other ensembling schemes) over multiple independent judgments of the same checklist. 
Third, \method can benefit from \emph{stronger judge models}, which provide more reliable evidence grounding and more consistent binary decisions. Beyond checklist-specific knobs, standard scaling strategies also apply, including using a \emph{stronger base model} and a \emph{larger group size} for advantage estimation, both of which typically improve optimization stability.

We expect these scaling directions to further stabilize training by suppressing residual stochasticity in tool use and judging. With sufficiently low-noise rewards, finer-grained reward assignment (e.g., step-level) may become viable, potentially retaining its fast early learning while avoiding premature collapse.

\section{Conclusion}
\method presents a scalable reinforcement learning framework for multi-turn, multi-step tool-using agents by replacing verifiable rewards with checklist rewards, which is fine-grained, binary, evidence-grounded criteria that make LLM judging more stable and interpretable. By adopting a ``sparse in reward assignment, dense in evaluation criteria'' strategy and training within an LLM-simulated tool environment, \method improves over supervised fine-tuning across multiple benchmarks and shows stronger generalization to complex, long-horizon tool-use behaviors where verifiable rewards are unavailable.

%% file: sec_arxiv/8_appendix.tex
\newpage

\section*{\LARGE Appendix}

\subsection{Prompts}
We force models to generate output in JSON format to ensure instruction following and the output can be parsed.

\subsubsection{Prompt of Checklist Labeling}
\label{sec:prompt_checklist_annotation}
The prompt for Checklist Annotation is shown in Prompt~\ref{pr:prompt_checklist_annotation}. We use GPT-5 to label the Checklist. The parameter ``effort'' is set to ``high'' for high quality. Each trajectory only costs approximately \$0.1 on average, making it practical to scale checklist labeling to large datasets without significant overhead.

\subsubsection{Prompt of LLM-based Filtering}
\label{sec:prompt_llm_based_filtering}
The prompt for LLM-based Filtering is shown in Prompt~\ref{pr:prompt_LLM_based_filtering}
We use GPT-5~\citep{openai2025gpt5} as the filter model. To reduce API costs while maintaining high filtering quality, we adopt an aggressive, progressive evaluation strategy: each sample is sequentially evaluated twice at \texttt{low} effort, twice at \texttt{medium} effort, and twice at \texttt{high} effort. If any single evaluation flags the sample as problematic, it is immediately discarded without further processing. This aggressive early-exit mechanism ensures that only high-confidence, high-quality samples survive the filtering pipeline, at the cost of potentially discarding some borderline cases.

\subsubsection{Prompt of CoT Compression}
\label{sec:prompt_shorten_thinking}
The prompt for CoT Compression is shown in Prompt~\ref{pr:prompt_shorten_thinking}. We use the default setting of GPT-5

\subsubsection{Prompt of LLM Judge}
\label{sec:prompt_llm_judge}
The prompt for LLM judge is shown in Prompt~\ref{pr:prompt_llm_judge}

\subsubsection{Prompt of Tool Simulation}
\label{sec:prompt_tool_simulation}
The prompt for LLM judge is shown in Prompt~\ref{pr:prompt_tool_simulation}

\subsection{Cold Start hyper-parameters}
\label{app:cold_start_hyperparameters}
For cold-start training, we utilize the \textbf{LLaMAFactory} framework. The model is trained on the cold-start dataset for 2 epochs. We adopt the AdamW optimizer and a cosine learning rate schedule. The learning rate is set to \texttt{1e-6} with a warmup ratio of \texttt{0.1}. The batch size is \texttt{64}. The cold-start training is conducted on 8 H100 GPUs.

To better handle special tokens (e.g., \texttt{<think>}, \texttt{<tool\_call>}) that are not trained during pre-training, we explicitly initialize their embeddings using the average of semantically related tokens. For example, the embedding of \texttt{<think>} is initialized as the mean of the embeddings for \texttt{think} and \texttt{begin}. This stabilizes optimization and speeds up convergence. 

The initialization is as follows:
\begin{align*}
  \texttt{<think>} &\leftarrow \text{avg("think", "begin")} \\
  \texttt{</think>} &\leftarrow \text{avg("think", "finish")} \\
  \texttt{<tool\_call>} &\leftarrow \text{avg("tool", "call", "start")} \\
  \texttt{</tool\_call>} &\leftarrow \text{avg("tool", "call", "end")} \\
  \texttt{<im\_start>} &\leftarrow \text{avg("role", "enter")} \\
  \texttt{<im\_end>} &\leftarrow \text{avg("role", "exit")}
\end{align*}

The training loss curve is shown in Figure~\ref{fig:cold_start_loss_curve}

\subsection{RL Training Details}
\label{app:rl_training_details}
For reinforcement learning, we set the mini-batch size to 128 and the learning rate to 3e-6. The KL divergence loss coefficient is set to 0.001, and we sample 24 or 48 trajectories for one question as a group size. We adopt GRPO as our RL algorithm with the standard deviation term in the denominator set to 1, following~\citep{feng2025group}. This improves the stability of the policy updates during training as we use a larger group size to ensure that later turns also receive a sufficient number of samples for sampling. We set the group number of 48 and use trajectory level reward for our final \method model.

\subsection{Rule-based Filtering Criteria}
\label{app:rule_base_filtering_criteria}
The criteria include: (1) violations of tool schemas; (2) incorrect role ordering; (3) mismatches between tool calls and subsequent responses; (4) tool responses erroneously placed within assistant messages; (5) invalid JSON formatting; (6) duplicate tool schemas or names; and (7) missing or redundant thinking tags (\texttt{<think>}).

\newpage

\begin{tcolorbox}[breakable,colback=promptbg,colframe=black!20,boxrule=0.5pt,arc=2pt,title=Prompt for Checklist Labeling]
\label{pr:prompt_checklist_annotation}
\begin{Verbatim}
# Instruction

You are an evaluation designer for **multi-turn, multi-step tool-use** dialogues.

## Your Task

Given a reference message list containing user, assistant, and tool steps, **produce a concise, per-turn checklist** of binary, observable criteria for judgment.
The checklist is used to judge whether another assistant meets the user's requirements.
One checklist per turn.

## Target of the assistant
The assistant needs to resolve the user's query in each turn.
It must analyze the user's intent in the private thinking, use tools to gather new information if necessary, plan the next steps based on the updated information and provide an user-visible reply to user.

## Input Format

### Conversation structure (multi-turn, multi-step)

* The conversation is chronological and split into **turns**.
* In each **turn**, there may be several steps from user, assistant, and tool:
  1. The **user** message appears **once** with questions or requirements.
  2. The **assistant** may think privately (Note: assistant content includes private thinking between <think> and </think>) and then either:
     * call one single tool or call multiple tools, **or**
     * generate a user-visible reply directly without calling tools.
  3. **Tool** messages return results to the preceding assistant message with tool calls.
  4. Repeat steps 2 and 3 until the turn ends.
* A turn **ends** when the assistant produces a user-visible reply after thinking.
* Only the **user-visible reply** is seen by the user.

### Candidate tools

You will also be given the schema of candidate tools for conversation. The tool calling should follow the schema (function name, required parameters, type of parameter)

### Message JSON schema (per step)

```json
{
  "role": "user|assistant|tool",
  "turn": 0,
  "step": 0,
  "content": "string containing either hidden thinking, user-visible reply, or tool output",
  "tool_calls": [
    {
      "id": "",
      "type": "function",
      "function": {
        "name": "TOOL_NAME",
        "arguments": { "Param": "Value", "...": "..." }
      }
    }
  ] # or None and []
}
```

* `turn` indexes start at **0**; `step` indexes start at **0** within each turn.

## Rules for the Checklist

1. Each item must be a **YES/NO** question with an **objective pass condition**.
2. Items must be **observable** from user messages, assistant private thinking/tool calls/user-visible reply, and tool responses.
3. For each item, specify **evidence pointers** that reference specific assistant or tool step, not user at step 0.
4. If the task has prerequisite tool response (e.g., "search before analyze"), encode them via **`depends_on`**. The dependence must be a tool step.
5. Within a turn, the checklist should cover **key requirements** implied by that turn’s user request, tool usage, constraints, and final reply (correctness, comprehensiveness, no hallucination, constraints, formatting, key reasoning steps, etc.).
6. Keep items atomic: ensure each checklist item evaluates a single, independent condition without combining multiple actions or operations.
7. Avoid purely stylistic or format checks; focus on key step to solve the user's requirements.
8. The question should focus on a specific part of the response, such as assistant.tool_calls, assistant.content.thinking, assistant.content.user_visible_reply, or tool.content (focus_on).
9. Allow procedurally different operations, intermediate conclusions, or derived facts **as long as they produce the same verifiable result and strictly follow the user's requirements**.
10. Provide a **weight** for every item (0–1) and normalize weights so they **sum to 1.0 per turn**, reflecting each requirement’s contribution and necessity to the final user-visible reply.
11. For each item, include a must_pass_to_continue boolean. True means this item must pass; otherwise the conversation should not proceed to the next turn (critical failure). False means non-critical; failure is tolerable but counted against quality.
12. The reference messages may contain some failed attempts. The checklist should not mention anything about those unsuccessful attempts or self-correction.
13. Assume there is no error in tool calling.

### Supplementary rules
1. Do not limit the number of tool calling.
2. Determine whether the value must match exactly or if a certain tolerance is acceptable.
3. Determine whether the parameter of tool calling must match exactly or if a certain tolerance is acceptable
4. The question about tool should align with the schema of candidate tool, e.g., argument with default value is not necessary.
5. Do not make any assumptions in the question, e.g., using if or when is question.
6. turn and step index should not appear or be refered to in checklist focus_on, question, pass_condition or failure_examples.

## How the Checklist Will Be Used

We evaluate **every assistant step with possible following tool response steps** within a turn to determine which checklist items become newly satisfied **relative to the previous assistant step** (for `step=0` there is no previous step). We **do not** require the model to complete items at specific, pre-ordained steps from the input log; instead, we assess whether **all requirements for that turn** are satisfied **by the end of the turn**, regardless of which assistant step achieved them or how assistant achieved them.

## Examples

from should be one of user.content|assistant.tool_calls|assistant.content.thinking|assistant.content.user_visible_reply|tool.content
[
  {
    "turn": 0,
    "checklist": [
      {
        "id": "C0", # start from 0 in each turn
        "evidence": [{
          "turn": TURN_INDEX,
          "step": STEP_INDEX,
          "from": "...",
          "snippet": "..."
        }],
        "focus_on": "assistant.tool_calls",
        "question": "Did the assistant call the required tool TOOL_NAME with the correct parameter Param=Value?",
        "pass_condition": "There exists an assistant tool call with name=TOOL_NAME and arguments.Param == Value or similar value.",
        "failure_examples": [
          "No tool call observed",
          "Wrong parameter value"
        ],
        "required_for_next_turn": true,
      },
      {
        "id": "C1",
        "evidence": [{
          "turn": TURN_INDEX,
          "step": STEP_INDEX,
          "from": "...",
          "snippet": "..."
        }],
        "focus_on": "tool.content",
        "question": "Did the assistant get xxx by calling the tool TOOL_NAME?",
        "pass_condition": "The assistant gets xxx from the tool response",
        "failure_examples": [
          "No tool response observed",
          "Wrong information from the tool"
        ],
        "required_for_next_turn": true,
      },
      {
        "id": "C2",
        "evidence": [{
          "turn": TURN_INDEX,
          "step": STEP_INDEX,
          "from": "...",
          "snippet": "..."
        }],
        "focus_on": "assistant.content.user_visible_reply",
        "question": "Does the final user-visible answer mentioned xxx?",
        "pass_condition": "The assistant’s final reply content mentions xxx that answers user's question.",
        "failure_examples": [
          "Assistant does not mention xxx",
          "Numeric/text mismatch between answer and tool output"
        ],
        "required_for_next_turn": true,
      }
      // ...more items
    ],
    "dependence": {
      "C0": [], // if no dependence, use a empty list
      "C1": [],
      "C2": ["C1"] // dependence (e.g. C1 here) must focus on tool.content
    },
    "weight": {     
      "C0": 0.3,
      "C1": 0.3,
      "C2": 0.4
    }    // ... must match item weights and sum to 1.0
  },
  // ... next turn checklist
]
\end{Verbatim}
\end{tcolorbox}

\begin{tcolorbox}[breakable,colback=promptbg,colframe=black!20,boxrule=0.5pt,arc=2pt,title=Prompt for LLM-based Filtering]
\label{pr:prompt_LLM_based_filtering}
\begin{Verbatim}
This is a simulated set of messages among a user, an assistant, and tools. The tools listed are the candidate tools. The assistant will first think (inside <think> and </think>; this part will be removed in post-processing and not shown to the user, and it should not be considered when judging whether there is an error), then decide whether to call a tool or produce a final response.

Does the logic for tool calling by the assistant follow the user's query, have no ambiguities, and can realistically occur in real scenarios? Are there any mistakes or flaws? Is it something that could exist in reality?
If there is no problem, answer true; if there is a problem, answer false.
You should be very strict.

Response format:
{
"Reasoning": string,
"NoError": true or false
}
\end{Verbatim}
\end{tcolorbox}

\vspace{15pt}

\begin{tcolorbox}[breakable,colback=promptbg,colframe=black!20,boxrule=0.5pt,arc=2pt,title=Prompt for CoT Compression]
\label{pr:prompt_shorten_thinking}
\begin{Verbatim}
# Instruction
You are given a multi-turn multi-step conversation consisting of messages. Each message has:
- role: one of [system | user | assistant | tool]
- content: textual content 
- thinking (for assistant messages, there is a thinking section)
- tool_call / tool_result

# Task:
For every assistant message that contains a thinking section, produce a concise rewritten thinking section that preserves all essential reasoning, decisions, constraints, and references needed to justify the reply, while removing verbosity, filler, repetitions, speculative or unneeded self-talk.

# Important Requirements:
1. Preserve Meaning: Do not change conclusions, assumptions, selected tools, or rationale ordering unless reordering improves clarity without altering logic.
2. Keep Necessary Steps: Retain key logical steps, intermediate conditions, disambiguations, and any constraints that influence the final answer or tool choice.
3. Do NOT invent new facts or reasoning not present in the original thinking.
4. Do NOT shorten so aggressively that causal links or justification for tool calls become unclear.
5. Maintain references to tool names, parameters, or required outputs if they affect the final answer.
6. If the original thinking is already minimal, keep it (possibly with tiny clarity edits).
7. If a thinking section is empty or missing, output one for that item.
8. Output Format must be strict JSON as described below (no extra commentary).

# Input JSON Schema (example):
{
  "tools": [
    // tool schemas
  ]
  "messages": [
    {
      "role": "user",
      "content": "...",
      "tool_calls": []
    },
    {
      "role": "assistant",
      "content": {
        "thinking": "Reason for choosing tool ...",
        "reply": "(May be empty if just a tool call step)"
      }
      "tool_calls": [ ... ],
    },
    {
      "role": "tool",
      "content": "Tool result ...",
      "tool_calls": []
    },
    {
      "role": "assistant",
      "content": {
        "thinking": "LONG INTERNAL REASONING TEXT ...",
        "reply": "Visible answer to user ..."
      }
    }
  ]
}

# Output format:
Return a JSON array aligned with assistant messages order. Each element corresponds to one assistant message.
The total number should be the same.

[
  {
    "thinking": "..."
  },
  ...
]
\end{Verbatim}
\end{tcolorbox}

\newpage
\begin{tcolorbox}[breakable,colback=promptbg,colframe=black!20,boxrule=0.5pt,arc=2pt,title=Prompt for LLM-as-a-Judge]
\label{pr:prompt_llm_judge}
\begin{Verbatim}
# Role
You are a precise checklist evaluator. Your sole task is to judge whether the messages between user, assistant and tool satisfie the provided criteria.

# Objective
Produce a strict JSON verdict (no extra text) based on the instructions below.

# Criteria
**Question:** {this_turn_checklist['question']}
**Focus on:** {this_turn_checklist['focus_on']}
**Pass condition:** {this_turn_checklist['pass_condition']}
**Failure examples:** {this_turn_checklist['failure_examples']}
**Reference snippet:** {reference_snippet}

# Previous Messages
{{messages_str_before_this_turn}}
# Current Messages to Evaluate
{{messages_str_in_this_turn}}

# Special rule of tool call
If there is no tool call in tool_call part but there are some tool calls in content.thinking part, it means these tools' format are not correct and all tool calls are not valid.If there is error in tool response. The previous tool calls in latest assistant (only the latest one) are not valid.# Evaluation Process (Align each step to a JSON output field)
1. high_level_understanding_of_the_question:
   - Briefly restate what is being evaluated (the intent of the question + what compliance means here).
2. analysis_of_if_focus_on:
   - Check whether Focus on part presents in the Current Messages.
3. analysis_of_pass_condition:
   - Determine if the 'Pass condition' is fully satisfied.
4. analysis_of_failure_examples:
   - For EACH failure example pattern: state clearly 'triggered' or 'not triggered' with a brief justification.
5. answer:
   - Return true ONLY IF:
     * Focus on part is present.
     * The 'Pass condition' is fully met.
     * No failure example pattern is triggered.
   - Otherwise return false.

# Output Format
Return ONLY a single JSON object with exactly these keys:
{
  "high_level_understanding_of_the_question": str,
  "analysis_of_if_focus_on": str,
  "analysis_of_pass_condition": str,
  "analysis_of_failure_examples": str,
  "answer": bool
}
\end{Verbatim}
\end{tcolorbox}

\begin{tcolorbox}[breakable,colback=promptbg,colframe=black!20,boxrule=0.5pt,arc=2pt,title=Prompt for Tool Simulation]
\label{pr:prompt_tool_simulation}
\begin{Verbatim}
# SYSTEM PROMPT
You are a precise tool executor that learns from examples.
You will be given:
- Tool call JSON Schema
- Few-shot examples showing tool calls and their execution results
- A new tool call with specific arguments

Your task:
1) Learn the OUTPUT FORMAT from the provided examples - follow the exact structure, data types, and response patterns
2) Ensure FACTUAL CONSISTENCY - your output should align with the factual information demonstrated in the examples
3) For the new tool call:
   - Apply the learned format to the new arguments
   - Maintain factual consistency with example patterns
   - If arguments are similar to examples, adapt the example results appropriately
   - If arguments are significantly different, generate new results following the learned format and factual patterns
   - May need to fix some type or error in the examples
4) Handle errors gracefully - if arguments are invalid or missing, return error messages in the same format as examples

Critical constraints:
- Act as a silent function executor - NO explanations, suggestions, or hints
- NO guidance on how to fix errors or improve calls
- NO references to examples or comparisons
- Return ONLY the raw execution result as valid JSON
- For errors, return minimal error information without instructional content

Output requirements:
- First do some analysis on how to mock the execution results. Then return ONLY the execution result as valid JSON array or object
- No explanations, markdown, or code fences
- Follow the exact output structure learned from examples
- Maintain factual consistency with the example patterns
Format:
{
  "analysis": str,
  "execution_result": JSON array or object,
}

# USER PROMPT
{"\n".join(examples_lines)}

Current tool name: {tool.name}
Current tool input schema (JSON Schema):
{schema_str}
Current arguments (JSON):
{parameters}
Generate tool execution result in JSON format.
\end{Verbatim}
\end{tcolorbox}